\pdfoutput=1

\documentclass[11pt]{article}
\usepackage[final]{emnlp2021}
\usepackage{times}
\usepackage{latexsym}
\usepackage[T1]{fontenc}
\usepackage[utf8]{inputenc}
\usepackage{microtype}

\usepackage{graphicx}
\usepackage{subcaption}
\usepackage{booktabs} 
\usepackage{amsmath}
\usepackage{amssymb}
\usepackage{multirow}
\usepackage{array}

\newcommand{\para}[1]{\smallskip\noindent\textbf{#1}}

\newcolumntype{C}[1]{>{\centering\arraybackslash}p{#1}}

\author{ 
Ninareh Mehrabi\thanks{~ equal contribution}\quad  Pei Zhou\footnotemark[1]\quad
\\ \textbf{Fred Morstatter \quad 
Jay Pujara \quad 
Xiang Ren \quad Aram Galstyan}

\\
{University of Southern California and Information Sciences Institute} \\
{
\texttt{\{ninarehm,peiz,morstatt,jpujara,xiangren,galstyan\}@usc.edu}
}  
} 

\begin{document}
\title{\textit{Lawyers are Dishonest?} Quantifying Representational Harms in Commonsense Knowledge Resources}
\maketitle
\begin{abstract}

\emph{\textbf{Warning}: this paper contains content that may
be offensive or upsetting.}


Commonsense knowledge bases (CSKB) are increasingly used for various natural language processing tasks. Since CSKBs are mostly human-generated and may reflect societal biases, it is important to ensure that such biases are not conflated with the notion of commonsense. Here we focus on two widely used CSKBs, ConceptNet and GenericsKB, and establish the presence of bias in the form of two types of representational harms, \emph{overgeneralization} of polarized perceptions and representation \emph{disparity} across different demographic groups in both CSKBs. Next, we find similar representational harms for downstream models that use ConceptNet. Finally, we propose a filtering-based approach for mitigating such harms, and observe that our filtered-based approach can reduce the issues in both resources and models but leads to a performance drop, leaving room for future work to build fairer and stronger commonsense models.

\end{abstract}

\section{Introduction}
Commonsense knowledge is important for a wide range of natural language processing (NLP) tasks as a way to incorporate information about everyday situations necessary for human language understanding. Numerous models have included knowledge resources such as ConceptNet~\cite{speer2017conceptnet} for question answering~\cite{lin-etal-2019-kagnet}, sarcasm generation~\cite{chakrabarty-etal-2020-r}, and dialogue response generation~\cite{Zhou2018CCM,zhou-etal-2021-commonsense}, among others. 
However, commonsense knowledge resources are mostly human-generated, either crowdsourced from the public~\cite{speer2017conceptnet,sap2019atomic} or crawled from massive web corpora~\cite{bhakthavatsalam2020genericskb}. For example, ConceptNet originated from the Open Mind Common Sense project that collects commonsense statements online from web users~\cite{singh2002open}\footnote{ConceptNet also includes knowledge from expert-created sources such as WordNet~\cite{miller1995wordnet}} and GenericsKB consists of crawled text from public websites. One issue with this approach is that the crowdsourcing workers and web page writers may conflate their own prejudices with the notion of commonsense. 
For instance, we have found that querying for some target words such as ``\emph{church}'' as shown in Table~\ref{motivation_table} in ConceptNet, results in biased triples. 

\begin{table}
\scalebox{0.7}{
\centering
\begin{tabular}{p{1.8cm}|p{8.0cm}}
 \toprule
\textbf{Source}&\textbf{Examples}\\[0.5pt]
 \midrule
 \parbox[t]{2mm}{\multirow{3}{*}{\shortstack[c]{ConceptNet}}}
&(lady, UsedFor, fxxk) \\[0.5pt]
&(lawyer, RelatedTo, dishonest)\\[0.5pt]
&(church, UsedFor, brain washing)\\[0.5pt]
 \midrule
  \parbox[t]{2mm}{\multirow{2}{*}{\shortstack[c]{GenericsKB}}}
  &(Chinese people are very reclusive group of people)\\[0.5pt]
 &(Lawyers are registered menace to society)\\[0.5pt]
  \bottomrule
\end{tabular}
}
\caption{Biased cases in  ConceptNet and GenericsKB.  
}
\label{motivation_table}
\end{table}

The potentially biased nature of commonsense knowledge bases (CSKB), 
given their increasing popularity, raises the urgent need to quantify biases both in the knowledge resources and in the downstream models that use these resources.
We present the first study on measuring bias in two large CSKBs, namely ConceptNet~\cite{speer2017conceptnet}, the most widely used knowledge graph in commonsense reasoning tasks, and GenericsKB~\cite{bhakthavatsalam2020genericskb}, which expresses knowledge in the form of natural language sentences and has gained increasing usage.
We formalize a new quantification of ``representational harms,'' \textit{i.e.}, how social groups (referred to as ``\emph{targets}'') are perceived~\cite{barocas2017problem,blodgett-etal-2020-language} in the context of CSKBs. 

We consider two types of such harms in the context of CSKBs. One is \emph{intra-target overgeneralization}, indicating that ``\emph{common sense}'' in these resources may unfairly attribute a polarized (negative or positive) characteristic to all members of a target class such as ``\emph{lawyers are dishonest.}'' The other is \emph{inter-target disparity}, occurring when targets have significantly different coverage in the CSKB in terms of both the number of statements about the targets (e.g., ``\emph{Persian}'' might have much fewer CS statements than ``\emph{British}'') and perception toward the targets (``\emph{islam}'' might have more negative CS statements than ``\emph{christian}'').


We propose a quantification of \emph{overgeneralization} and \emph{disparity} in CSKBs using two proxy measures of polarized perceptions: sentiment and regard~\cite{sheng2019woman}. 
Applying the proposed metrics of bias to ConceptNet and GenericsKB, we find harmful overgeneralizations of both negative and positive perceptions over many target groups, indicating that human biases have been conflated with ``\emph{common sense}'' in these resources. We find severe disparities across the targets in demographic categories such as professions and genders, both in the number of statements and the polarized perceptions about the targets.

We then examine two generative downstream tasks and the corresponding models that use ConceptNet. Specifically, we focus on automatic knowledge graph construction and story generation and quantify biases in COMeT~\cite{bosselut2019comet} and CommonsenseStoryGen (CSG)~\cite{guan2020knowledge}.
We find that these models also contain the harmful overgeneralizations and disparities found in ConceptNet.
We then design a simple mitigation method that filters unwanted triples according to our measures in ConceptNet.
We retrain COMeT using filtered ConceptNet and show that our proposed mitigation approach helps in reducing both overgeneralization and disparity issues in the COMeT model but leads to a performance drop in terms of the quality of triples generated according to human evaluations. 
We open-source our data and prompts to evaluate biases in commonsense resources and models for future work \footnote{\url{https://github.com/Ninarehm/Commonsense\_bias}}.

 
\section{Quantifying Representational Harms}
\label{sec:formalization}

\subsection{Representational Harms}

Representational harms occur ``\emph{when systems reinforce the subordination of some groups along the lines of identity}'' and can be further categorized into stereotyping, recognition, denigration, and under-representation~\cite{barocas2017problem, crawford2017trouble,  blodgett-etal-2020-language}. 
This work aims to formalize representational harms specifically for \emph{a set of statements} about some target groups~\cite{nadeem2020stereoset}, e.g ``\emph{lawyer is related to dishonest}'' is a statement about the group ``\emph{lawyer.}'' 

When measuring such harms, we consider the core concept of \emph{polarized perceptions}: non-neutral views that can take the form of either prejudice that expresses negative views~\footnote{The word \emph{prejudice} is defined as ``\emph{preconceived (usually unfavorable) evaluation of another person}''~\cite{lindzey1968handbook}. We focus on only unfavorable prejudice and use favoritism for favorable evaluation.} or favoritism that expresses positive views toward a certain target perceived in the statement~\cite{10.1145/3457607}.

More formally, let $\mathbb{S} = \left \{  s_1, s_2, ..., s_n\right \}$ indicate the set of $n$ natural language statements $s_i$ and let $\mathbb{T} = \left \{  t_1, t_2, ..., t_m,\right \}$ indicate the set of $m$ targets such that each $t_j$ has appeared at least once in $\mathbb{S}$. Each statement $s_i$ contains a target $t_j$. We use $s_i^{+/-}(t_j)$ to indicate that the statement expresses a positive or negative perception toward the target $t_j$.
We are interested in quantifying the representational harms toward $\mathbb{T}$ in this set $\mathbb{S}$.

\subsection{Two Types of Harms}
To adapt the definition of representational harms to a sentence set, we define two sub-types of harms, \emph{intra-target overgeneralization} and \emph{inter-target disparity}, aiming to cover different categories of representational harms~\cite{barocas2017problem, crawford2017trouble}. We consider \emph{overgeneralization} that directly examines whether targets such as ``\emph{lawyer}'' or ``\emph{lady}'' are perceived positively or negatively in the statements (examples in Table~\ref{motivation_table}), covering categories including stereotyping, denigration, and favoritism. Then we consider \emph{disparity} across different targets in representation (do some targets have fewer associated statements and lower coverage) and polarized perceptions (whether some targets are more positively or negatively perceived).

\para{Intra-target Overgeneralization}
The ideal sentence set depicting a target group such as ``\emph{lawyer}'' or ``\emph{lady}'' should have neither \emph{favoritism} nor \emph{prejudice} toward the target group.
Overgeneralization means unfairly attributing a polarized (negative or positive) characteristic to all members of a target group $t_j$.
In the context of sentence sets $\mathbb{S}$, we define overgeneralization to be polarized sentences in the set regarding the targets. When the statement $s_i$ contains negatively-polarized views toward a target group such as ``\emph{lawyers are dishonest,}'' it demonstrates \textit{prejudice} toward lawyers. It is an overgeneralized statement because it implies that \emph{all} lawyers are not honest. The same logic applies to positively-polarized statements such as ``\emph{British people are brilliant}'' that constitute favoritism.

To measure the polarization in a sentence $s_i$, we use two approximations: sentiment polarity and regard~\cite{sheng2019woman}. We define an intra-target overgeneralization score using these measures for every target in $\mathbb{S}$. Formally, for each target $t_j$, we collect all statements in $\mathbb{S}$ that contain target $t_j$ to form a target-specific set $\mathbb{S}_{t_j}$, we then apply the sentiment and regard classifiers individually on all statements in $\mathbb{S}_{t_j}$ and report the average percentages of statements that are classified as positive or negative sentiment or regard labels for the target $t_j$. The negative and positive overgeneralization bias (prejudice and favoritism) for $t_j$ is quantified as:
\begin{equation} \label{eq:1}
     O^{-}(\mathbb{S}, t_j) = |\mathbb{S}^{-}_{t_j}||\mathbb{S}_{t_j}|^{-1} \times 100,
\end{equation}
\begin{equation} \label{eq:2}
     O^{+}(\mathbb{S}, t_j) = |\mathbb{S}^{+}_{t_j}||\mathbb{S}_{t_j}|^{-1} \times 100,
\end{equation}
where $|\mathbb{S}^{+/-}_{t_j}|$ is the number of statements with positive/negative polarization measured by sentiment or regard for the target $t_j$, i.e. $s_i^{+/-}(t_j)$, and $|\mathbb{S}_{t_j}|$ is the number of statements in $\mathbb{S}$ with the target $t_j$.

\para{Inter-target Disparity}
In addition to overgeneralized non-neutral views for each target group, we also study \textit{inter-target disparity} -- \textit{i.e.}, how different a target $t_j$ is perceived in the set $\mathbb{S}$ compared to other targets. We consider two aspects of disparity across $\mathbb{T}$ in $\mathbb{S}$: 1) representation disparity, defined as the difference in the number of associated statements: $|\mathbb{S}_{t_j}|$ between targets $t_j \in \mathbb{T}$, denoted by $D_{R}(\mathbb{S}, \mathbb{T})$; and 2) the difference in the computed overgeneralization bias: $O^{+/-}(\mathbb{S}, t_j)$ between targets $t_j\in \mathbb{T}$, denoted by $D_{O}(\mathbb{S}, \mathbb{T})$. Note that compared to $O^{+/-}(\mathbb{S}, t_j)$, $D_{O}(\mathbb{S}, \mathbb{T})$ is calculated over the full population of targets, thus measuring \emph{inter-target} disparity.
For both aspects, we measure disparity using variance as follows:
\begin{equation} \label{eq:3}
      D_{R}(\mathbb{S}, \mathbb{T}) =\mathbb{E}[(|\mathbb{S}_{t_j}|-|\overline{\mathbb{S}_{t}}|)^2],
\end{equation}
\begin{equation} \label{eq:4}
 D_{O}^{+/-}(\mathbb{S}, \mathbb{T}) =\mathbb{E}[(O^{+/-}(\mathbb{S}, t_j)-\overline{O^{+/-}(\mathbb{S}, t_j)})^2],
\end{equation}
where $|\overline{\mathbb{S}_{t}}|$ indicates the average number of statements for targets in $\mathbb{T}$ and $\overline{O^{+/-}(\mathbb{S}, t_j)}$ is the average overgeneralization bias for targets, ``+'' for favoritism and ``-'' for prejudice. The expectation $\mathbb{E}$ is taken over all targets $t_j \in \mathbb{T}$.

\subsection{Measuring Polarized Perceptions}\label{measure_quality}

Prior work~\cite{sheng2019woman} demonstrated that sentiment and regard are effective measures of bias (polarized views toward a target group).
Although this is still an active area of research, for now, these are promising proxies that many works in ethical NLP also have used to measure bias (e.g.~\citet{sheng2019woman, li2020unqover, NEURIPS2020_1457c0d6, sheng-etal-2020-towards,dhamala2021bold}). However, we acknowledge that there still exist problems with these measures as proxies for measuring bias and acknowledge the existence of noisy labels using these measures as proxies. To put this into test and to show that these measures can still be reliable proxies despite the aforementioned problems, we perform studies both including human evaluators in the loop as well as comparison of these measures with a keyword-based approach in this section. 

In order to determine the polarization of perception associated to a statement toward a group, we apply sentiment and regard classifiers on the statement containing the target group and obtain the corresponding labels from each of the classifiers. We then categorize the statement into favoritism, prejudice, or neutral based on the positive, negative, or neutral labels obtained from each of the classifiers. 

\para{Crowdsourcing Human Labels}
To validate the quality of these polarity proxies, we conduct crowdsourcing to solicit human labels on the statement polarity.
We asked Amazon Mechanical Turk workers to label provided knowledge from GenericsKB~\cite{bhakthavatsalam2020genericskb} and ConceptNet~\cite{speer2017conceptnet} with regards to favoritism, prejudice, and neutral toward a target group. 3,000 instances were labeled from ConceptNet and more than 1,500 from GenericsKB. The inter-annotator agreement in terms of Fleiss' kappa scores \cite{fleiss1971measuring} for this task was 0.5007 and 0.3827 for GenericsKB and ConceptNet respectively. 

\begin{table}[tb]
\scalebox{0.73}{
    \begin{tabular}{c | c c  | c c }
        \toprule
        CSKB & \multicolumn{2}{c}{GenericsKB} & \multicolumn{2}{c}{ConceptNet} \\
        \midrule
        Measure & Sentiment & Regard   & Sentiment & Regard \\
        \midrule
        Human Agreement &70.3\% & 60.9\% & 83.1\% & 75.4\%\\
        \bottomrule
    \end{tabular}}
    \caption{Agreement of sentiment and regard labels with human annotators in terms of accuracy.}
    \label{human_vs_re_sent_generics}
\end{table}

\begin{table}[tb]
\centering
\scalebox{0.72}{
\begin{tabular}{p{1.8cm}|ccc}
 \toprule
\textbf{CSKB}&\textbf{Method}&\textbf{Favoritism R/P/F1}&\textbf{Prejudice R/P/F1}\\[0.5pt]
 \midrule
 \parbox[t]{2mm}{\multirow{3}{*}{\shortstack[c]{GenericsKB}}}
&Regard&\textbf{0.551}/0.579/\textbf{0.565}&\textbf{0.809}/0.333/0.472 \\[0.5pt]
&Sentiment&0.441/0.622/0.516&0.432/\textbf{0.541}/\textbf{0.480}\\[0.5pt]
&Keyword&0.268/\textbf{0.643}/0.379&0.276/0.539/0.365\\[0.5pt]
 \midrule
  \parbox[t]{2mm}{\multirow{2}{*}{\shortstack[c]{ConceptNet}}}
  &Regard&\textbf{0.436}/0.383/0.408&\textbf{0.698}/0.342/\textbf{0.459}\\[0.5pt]
  &Sentiment&0.378/0.528/\textbf{0.440}&0.264/\textbf{0.531}/0.353\\[0.5pt]
 &Keyword&0.201/\textbf{0.556}/0.295&0.105/0.470/0.172\\[0.5pt]
  \bottomrule
\end{tabular}
}

\caption{Comparison of sentiment, regard, and baseline keyword-based approach in terms of favoritism and prejudice recall/precision/F1 scores.}
\label{recall}
\end{table}
\para{Alignment with Human Labels}
We compare human labels with those obtained from sentiment and regard classifiers to check the validity of these measures as proxies for overgeneralization. As shown in Table \ref{human_vs_re_sent_generics}, we found reasonable agreement in terms of accuracy for sentiment and regard with human labels. This was also confirmed in previous work \cite{sheng2019woman} in which sentiment and regard were shown to be good proxies to measure bias. 

\para{Comparison with Keyword-based Approach}
We also compare the sentiment and regard classifiers to a keyword-based baseline, in which we collect a list of biased words that could represent favoritism and prejudice from LIWC \cite{doi:10.1177/0261927X09351676} and Empath \cite{10.1145/2858036.2858535}. 
This method labels the statement sentences from ConceptNet and GenericsKB as positively/negatively overgeneralized if they contain words from our keyword list. As shown in Table \ref{recall}, this method has a significantly lower recall and overall F1 value in identifying favoritism and prejudice compared to sentiment and regard measures.


\section{Representational Harms in CSKBs} \label{sec:3}
\begin{figure*}[h]
\centering
\begin{subfigure}[b]{0.24\textwidth}
\includegraphics[width=\textwidth,trim=0cm 0cm 0cm 0cm,clip=true]{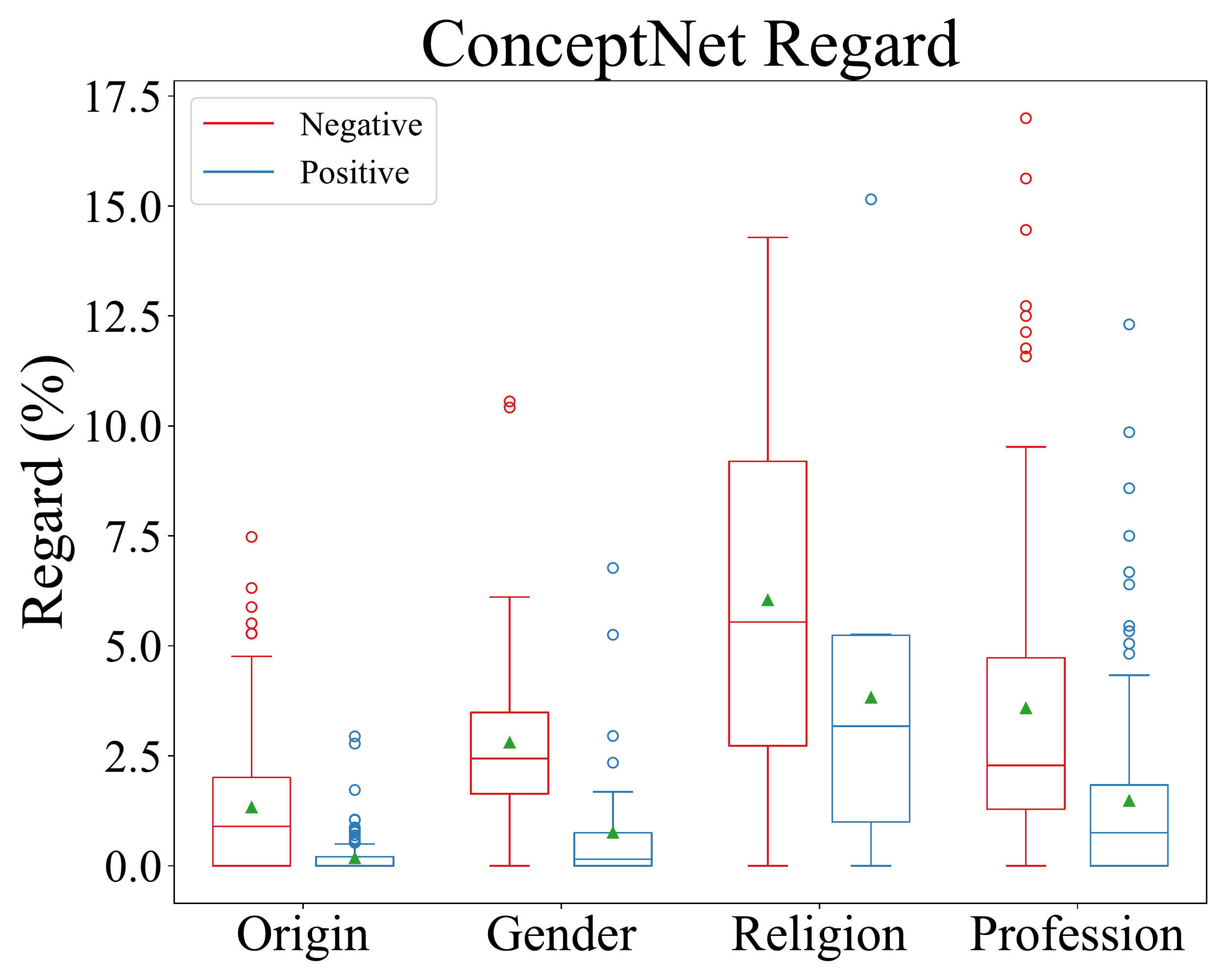}
\end{subfigure}
\begin{subfigure}[b]{0.24\textwidth}
\includegraphics[width=\textwidth,trim=0cm 0cm 0cm 0cm,clip=true]{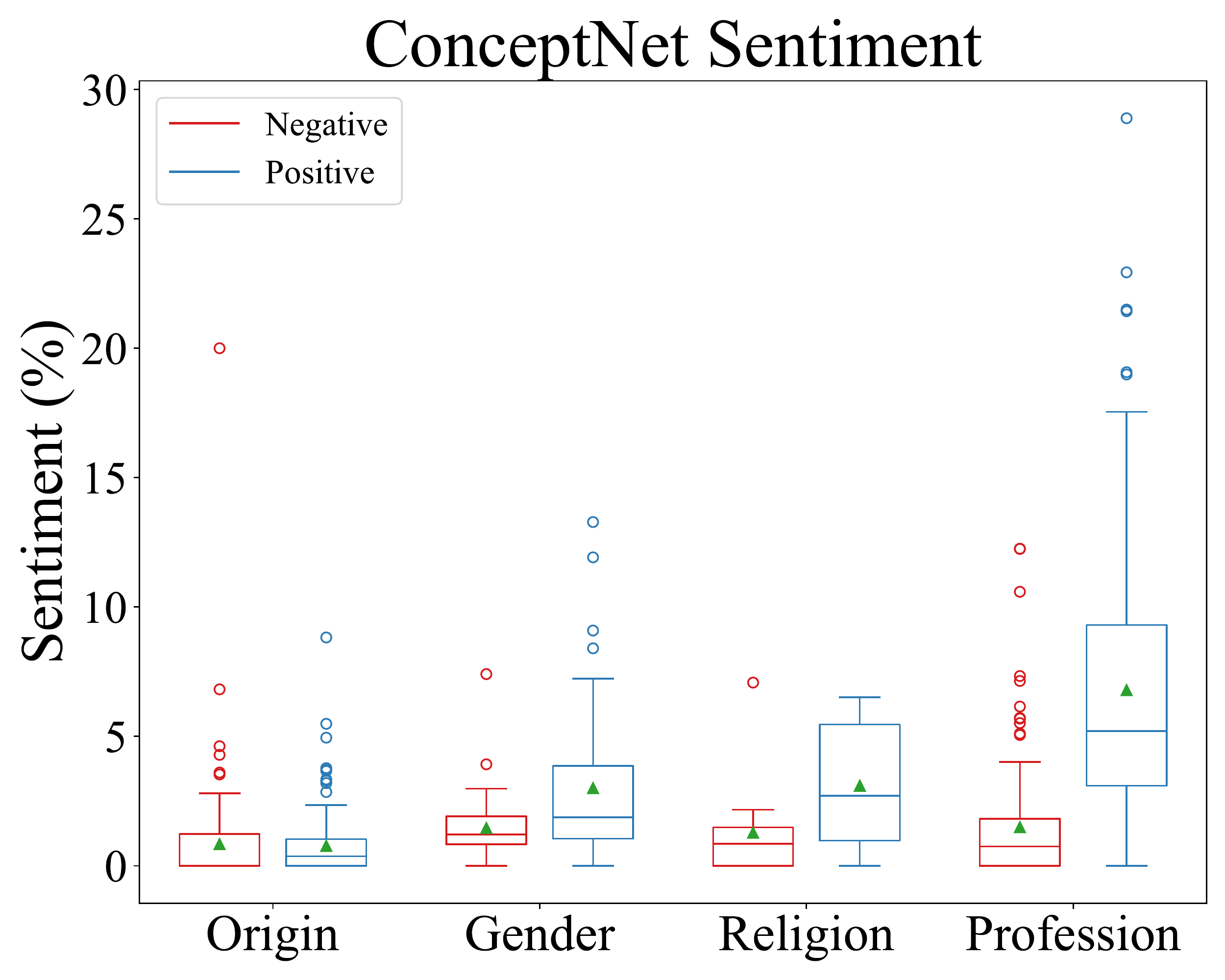}
\end{subfigure}
\begin{subfigure}[b]{0.24\textwidth}
\includegraphics[width=\textwidth]{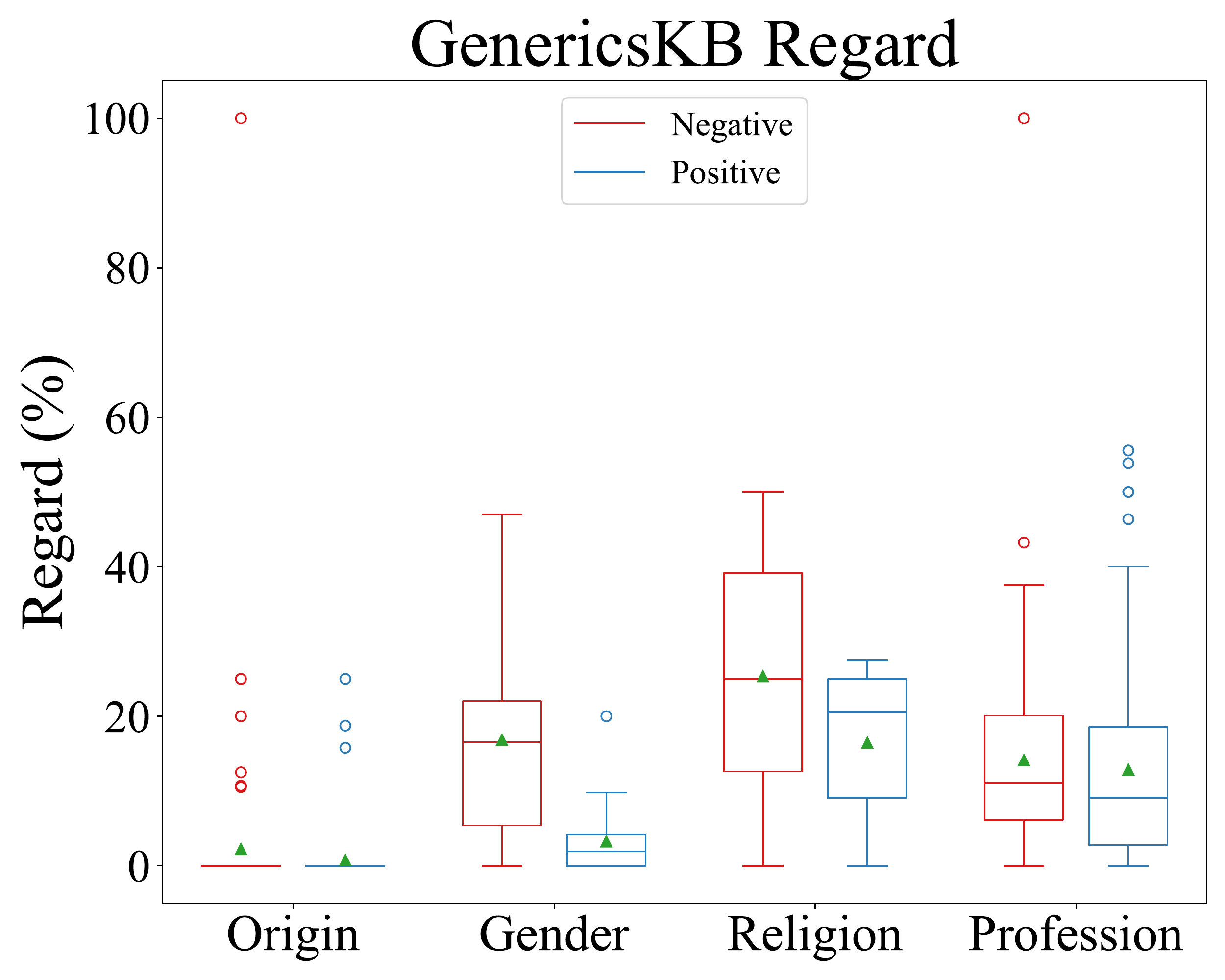}
\end{subfigure}
\begin{subfigure}[b]{0.24\textwidth}
\includegraphics[width=\textwidth]{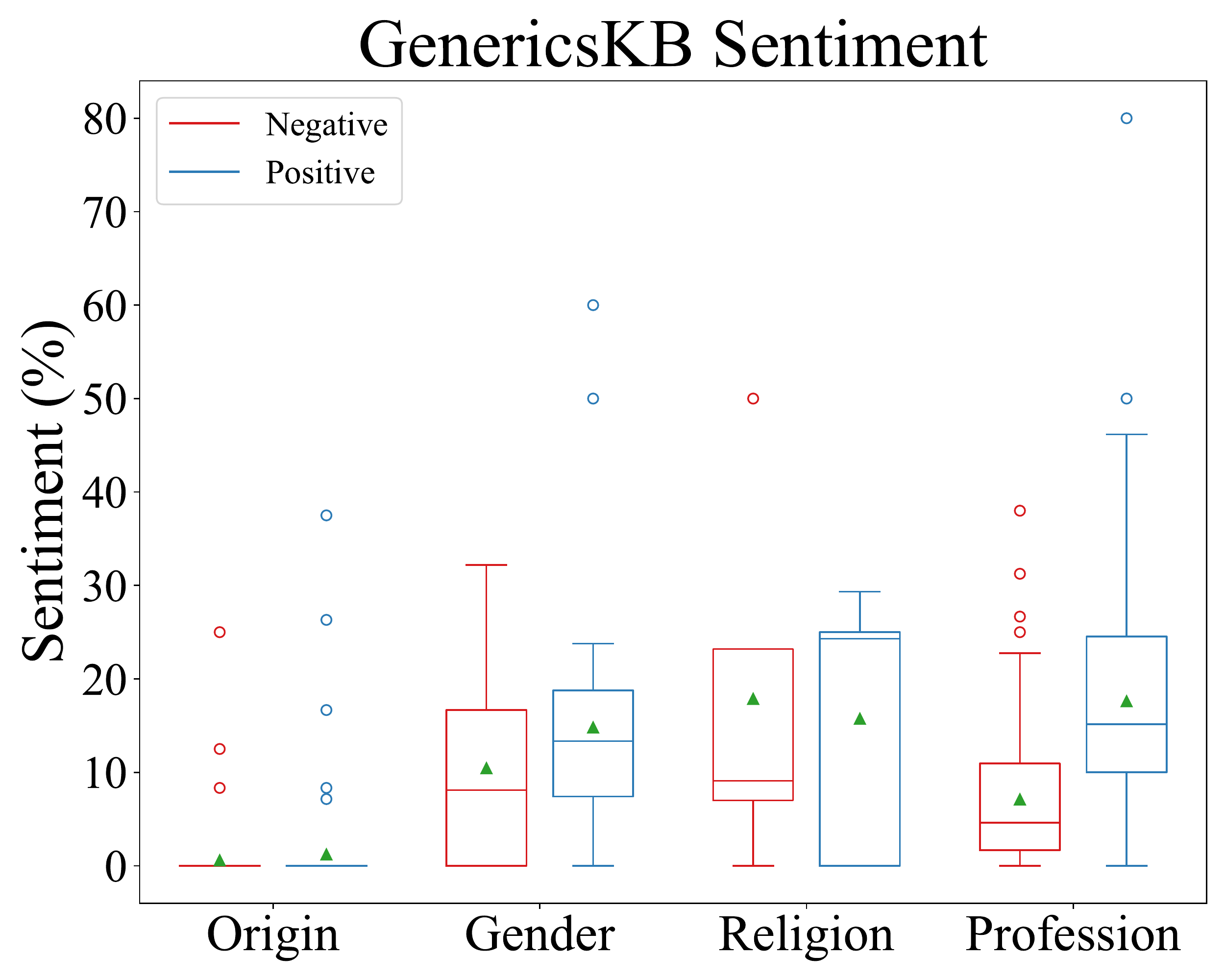}
\end{subfigure}

\caption{Negative and positive regard and sentiment results from ConceptNet and GenericsKB. We find outlier target groups with high regard and sentiment percentages that show the severity of \textbf{overgeneralization} issues. We also find large \textbf{variation/disparity} in the number of negative or positive triples for groups in the same category indicated by the span of boxes.
}
\label{fig:Conceptnet_Bias_results}
\end{figure*}
\begin{figure*}[h]
\begin{subfigure}[b]{0.33\textwidth}
\includegraphics[width=\textwidth,trim=7.3cm 0cm 7.3cm 0cm,clip=true]{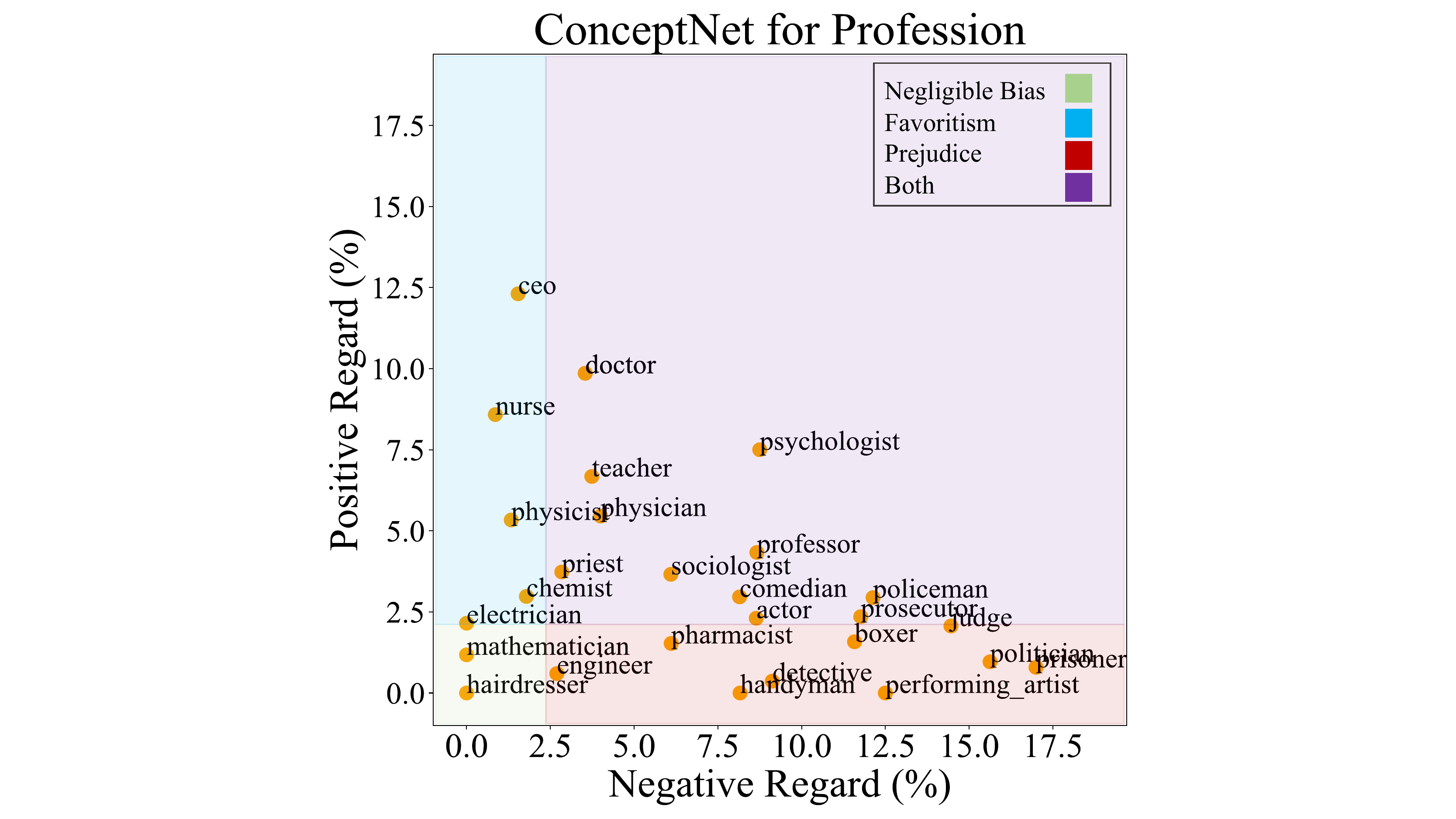}
\end{subfigure}
\begin{subfigure}[b]{0.33\textwidth}
\includegraphics[width=\textwidth,trim=7.3cm 0cm 7.3cm 0cm,clip=true]{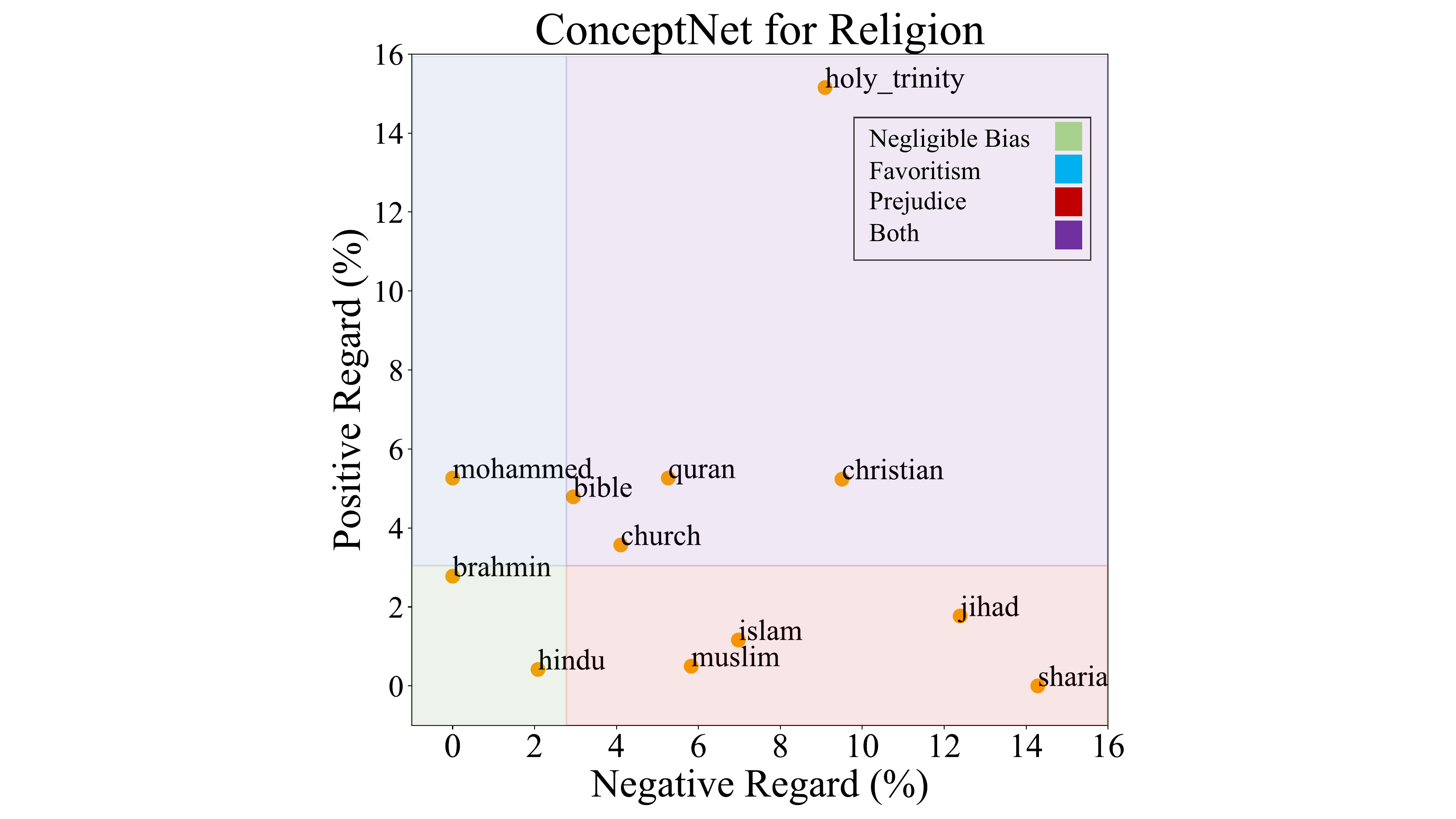}
\end{subfigure}
\begin{subfigure}[b]{0.33\textwidth}
\includegraphics[width=\textwidth,trim=8.3cm 0cm 6.3cm 0cm,clip=true]{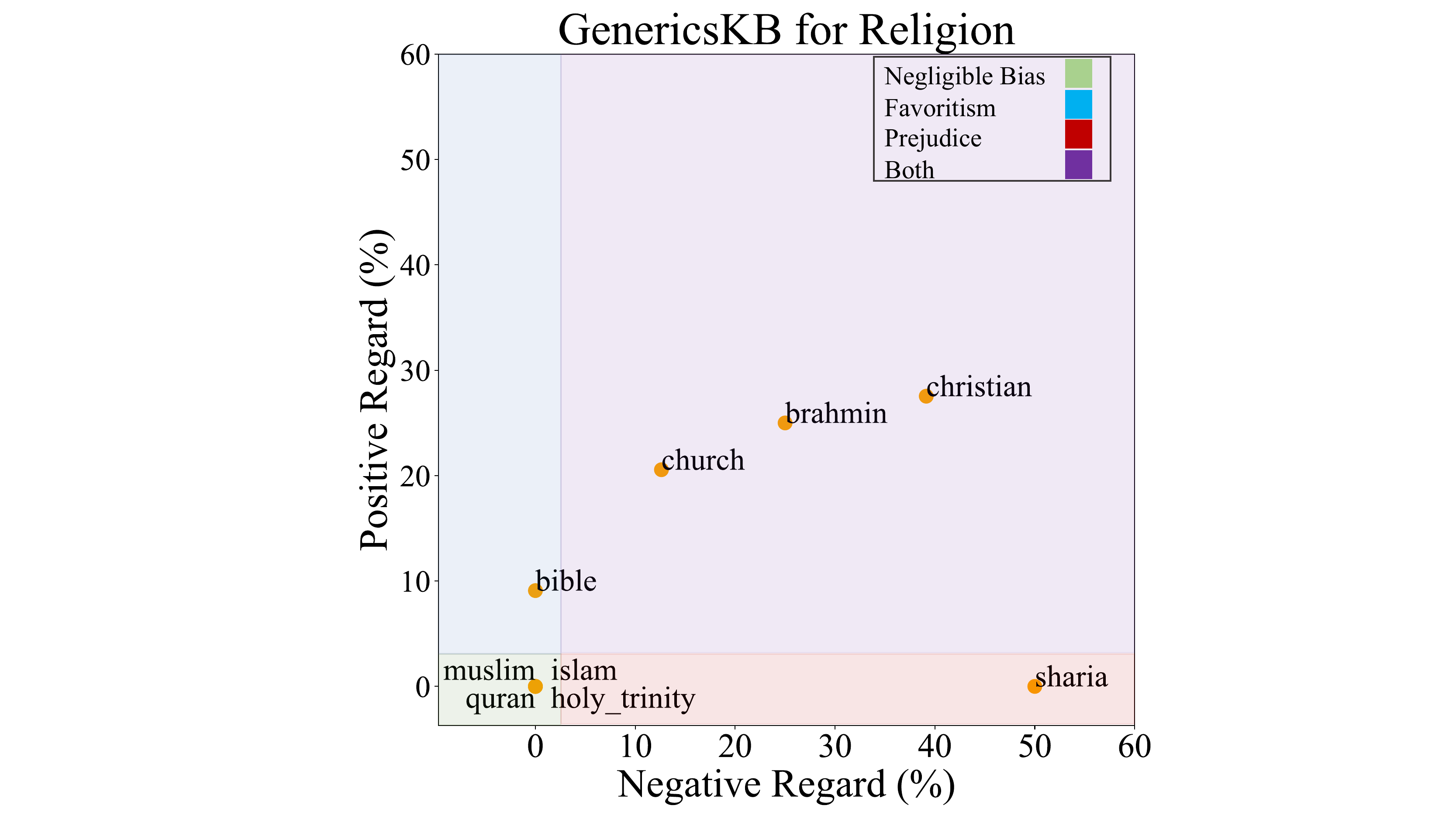}
\end{subfigure}
\caption{Examples of targets from the ``\emph{Profession}'' and ``\emph{Religion}'' categories from~\citet{nadeem2020stereoset} labeled by the regard measure. Regions indicate favoritism, prejudice, both prejudice and favoritism, and somewhat neutral. Higher negative regard percentages indicate prejudice-leaning and higher positive regard percentages indicate favoritism-leaning. We also compare ConceptNet~\cite{speer2017conceptnet} and GenericsKB~\cite{bhakthavatsalam2020genericskb} on the ``\emph{Religion}'' category and find similar polarized perceptions of certain groups, despite a much larger percentage range for GenericsKB.}
\label{fig:scatter_prof_sent}

\end{figure*}

\begin{figure*}[h]
\centering
\begin{subfigure}[b]{0.49\textwidth}
\includegraphics[width=\textwidth,trim=2cm 9.2cm 9cm 0cm,clip=true]{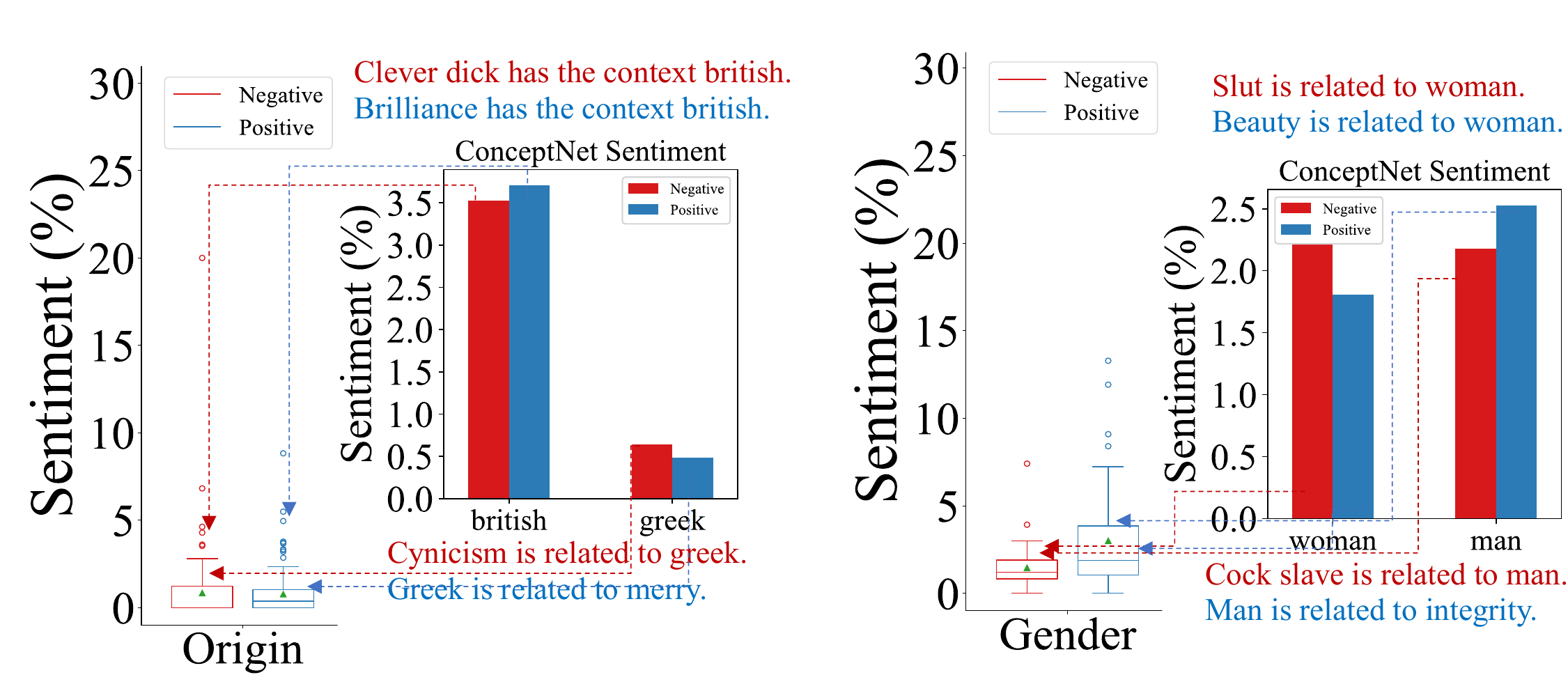}
\end{subfigure}
\begin{subfigure}[b]{0.49\textwidth}
\includegraphics[width=\textwidth,trim=2cm 9cm 7.5cm 0cm,clip=true]{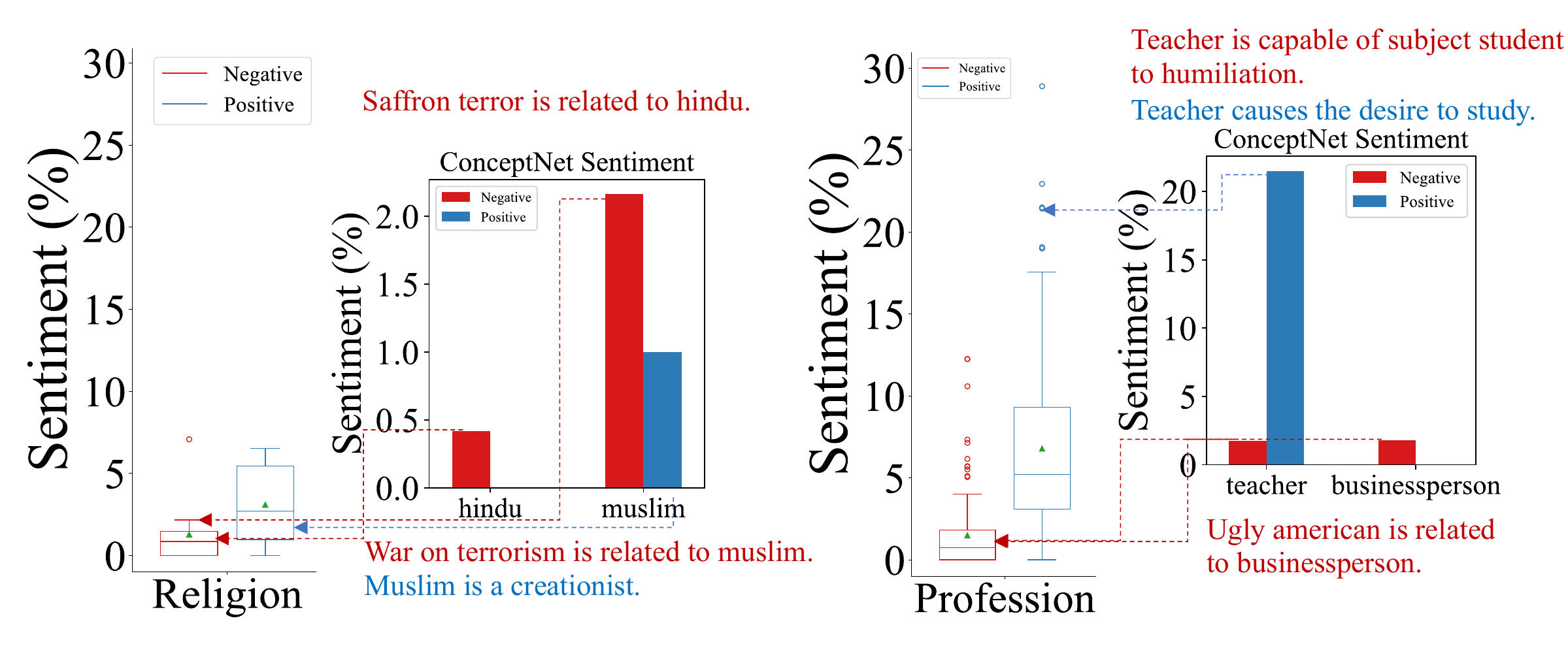}
\end{subfigure}
\caption{Four different representations from four categories each demonstrating a certain aspect of bias. In ``\emph{Origin}'' category, we can observe extreme overgeneralization toward ``\emph{british},'' in ``\emph{Gender}'' category both target groups are overgeneralized, in ``\emph{Religion}'' extreme prejudice toward ``\emph{muslim},'' and in ``\emph{Profession}'' extreme favoritism toward ``\emph{teacher}'' target group. Each case is accompanied with an example of negative and positive associations detected by sentiment.
}
\label{fig:sample_Bias_results}
\end{figure*}

\begin{figure}[h]
\centering
\begin{subfigure}[b]{0.24\textwidth}
\includegraphics[width=\textwidth,trim=0.2cm 0.2cm 0.2cm 0.2cm,clip=true]{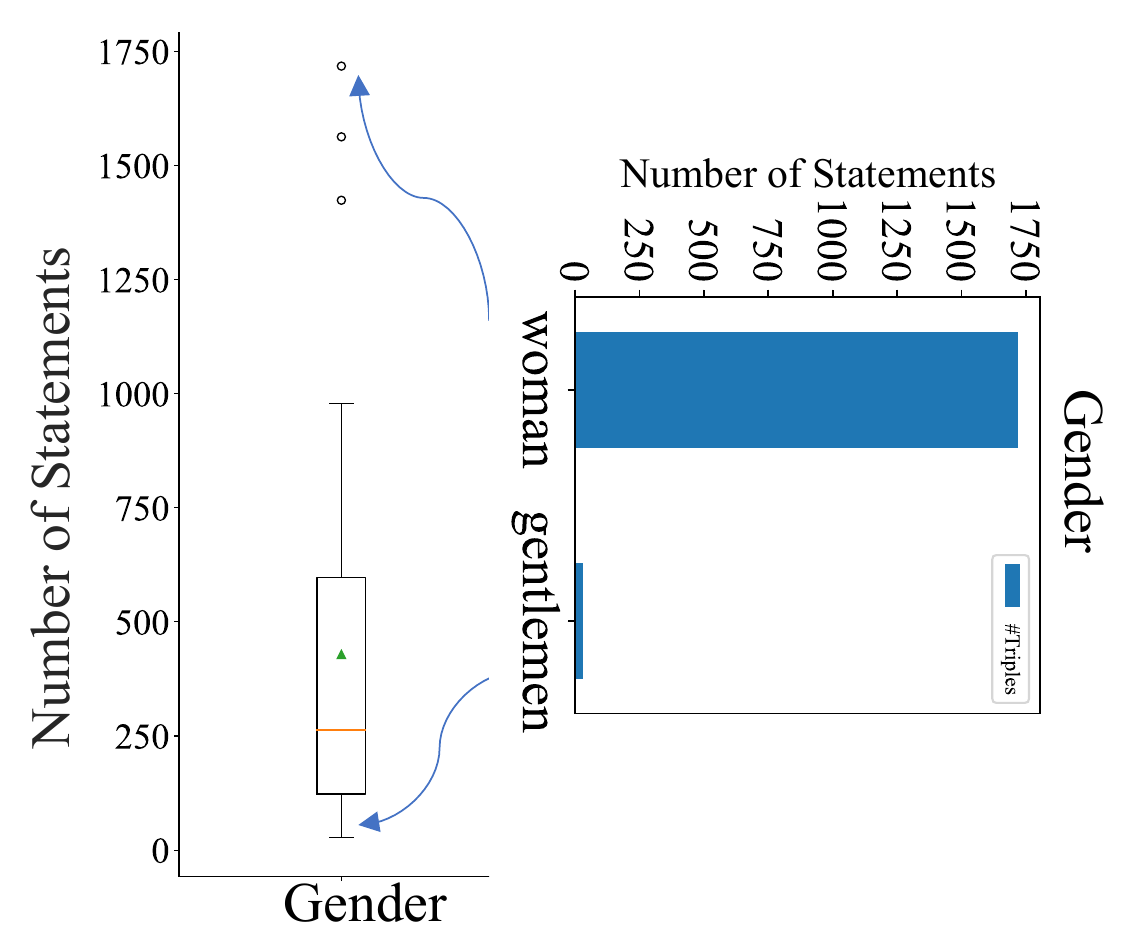}
\caption{ConceptNet Gender}
\end{subfigure}
\begin{subfigure}[b]{0.22\textwidth}
\includegraphics[width=\textwidth,height=0.13\textheight,trim=0.2cm 0.2cm 0.2cm 0.2cm,clip=true]{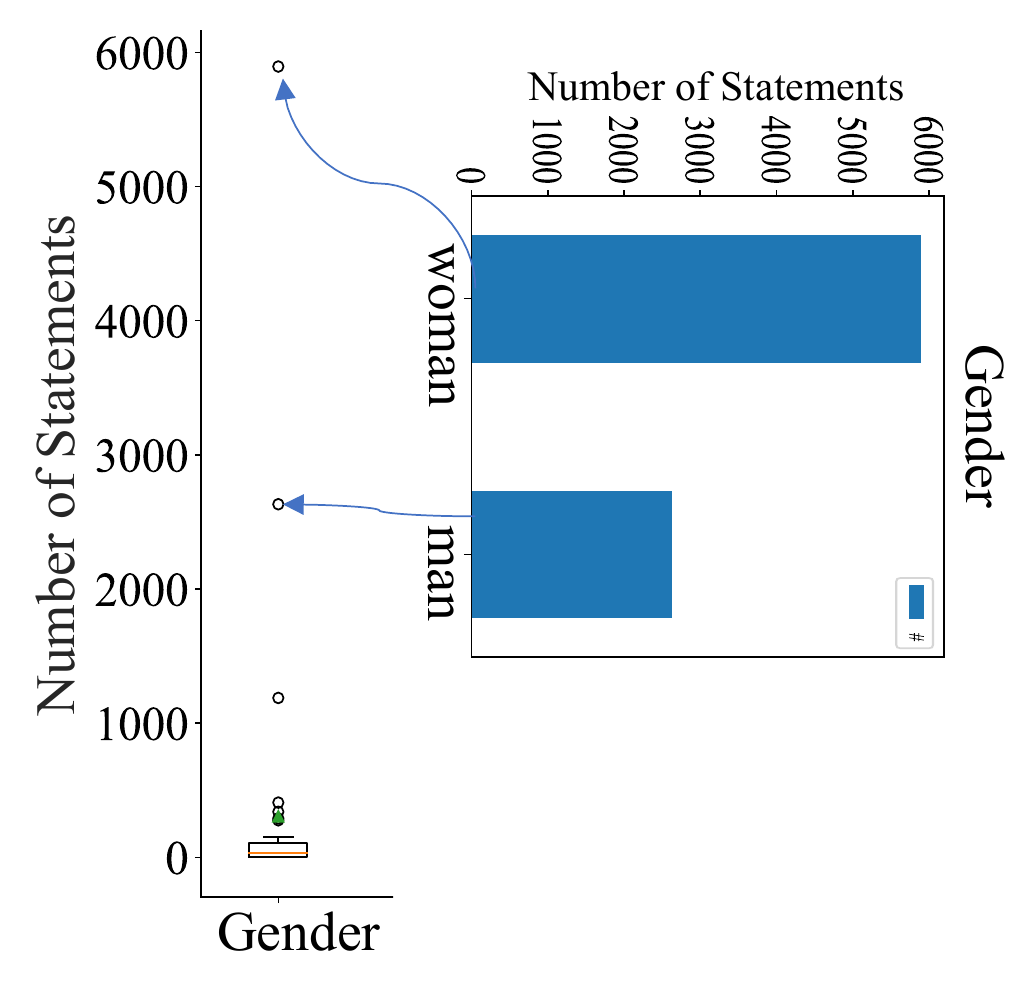}
\caption{GenericsKB Gender}
\end{subfigure}
\begin{subfigure}[b]{0.23\textwidth}
\includegraphics[width=\textwidth,trim=0.2cm 0.2cm 0.2cm 0.2cm,clip=true]{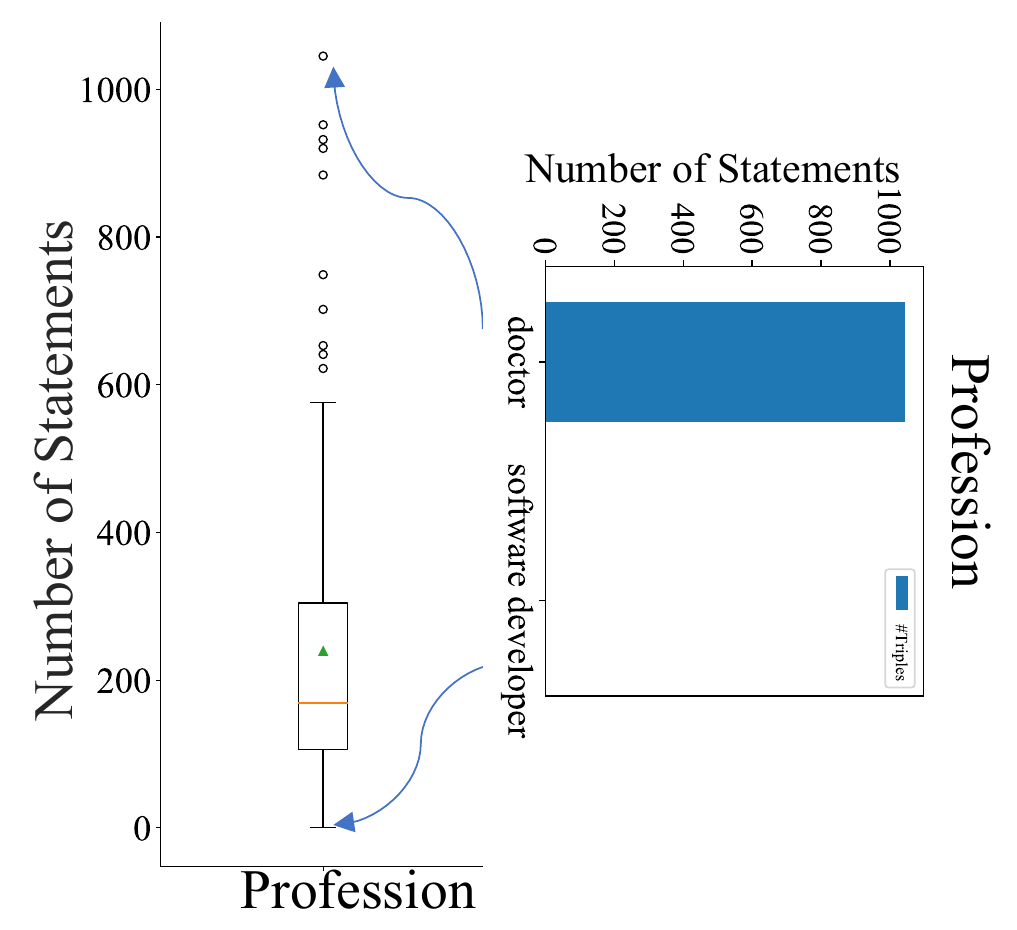}
\caption{ConceptNet Profession}
\end{subfigure}
\begin{subfigure}[b]{0.23\textwidth}
\includegraphics[width=\textwidth,trim=0.2cm 0.4cm 0.2cm 0.2cm,clip=true]{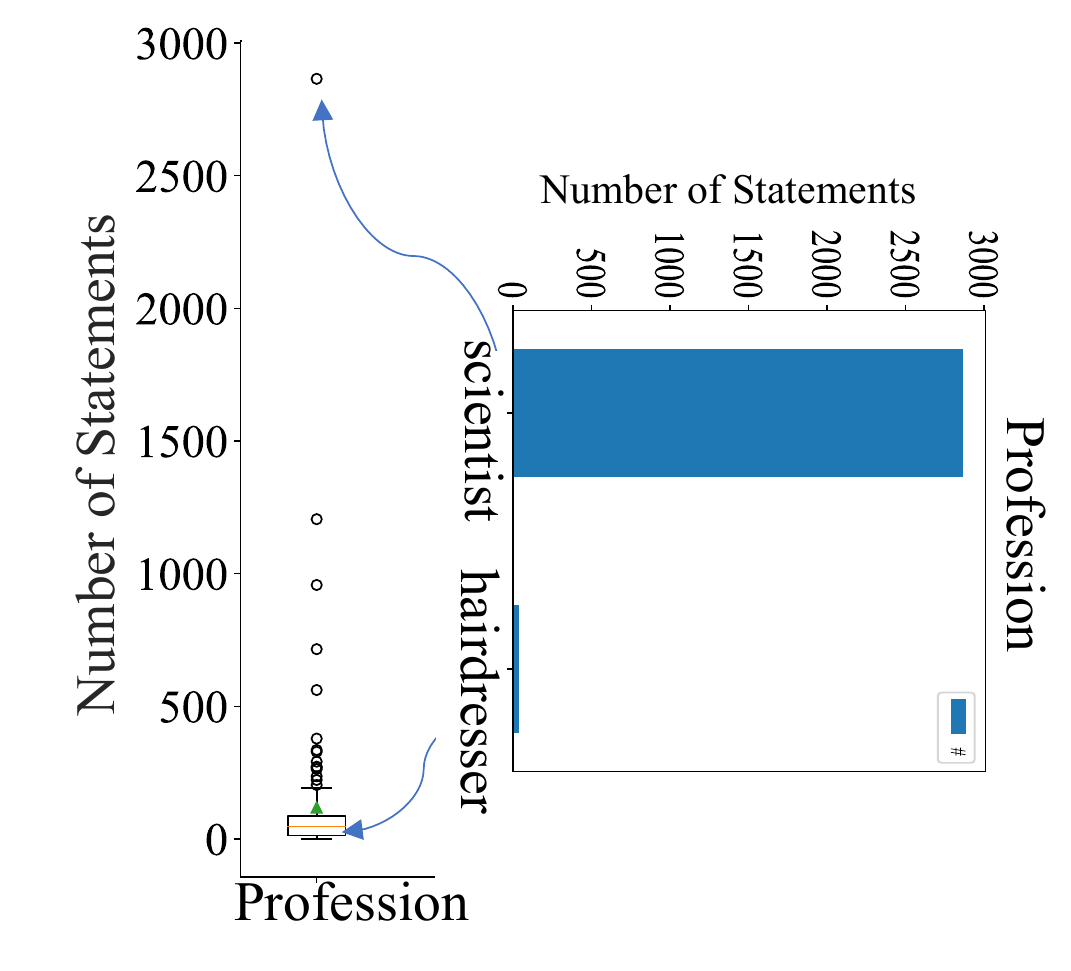}
\caption{GenericsKB Profession}
\end{subfigure}
\caption{Box plots demonstrating the \textbf{representation disparity} in terms of number of triples/sentences for ``\emph{Gender}'' and ``\emph{Profession}'' categories from ConceptNet and GenericsKB. We find similarly severe disparities in two KBs with the number of sentences ranging much more for GenericsKB compared to ConceptNet.}
\label{fig:sample_freq_results}
\end{figure}

Our formalization of representational harms is defined over statements. GenericsKB uses naturally occurring sentences, so our measures (in Sec.~\ref{sec:formalization}) directly apply. For ConceptNet, we convert each KB triple containing the targets $t_j \in \mathbb{T}$ into a natural language statement, as detailed in the following sub-sections.

\subsection{Data Preparation}

\noindent
\textbf{Selection of Target Groups}
Let $G$ be the graph of the CSKB that consists of commonsense assertions in the form of triples $(s,r,o)$ representing ``\emph{subject-relation-object}''. Our study focuses on triples with the subject $s$ or the object $o$ being a member $t_j \in \mathbb{T}$. 
We first collect a list of targets $t_j$ from the StereoSet dataset~\cite{nadeem2020stereoset}. The collected targets are organized within 4 different categories: \textit{origin}, \textit{gender}, \textit{religion}, and \textit{profession}. We renamed the ``\emph{race}'' category from ~\cite{nadeem2020stereoset} to ``\emph{origin}'' to be more precise as words such as \textit{British} may not necessarily represent a race but more of the origin or nationality of a person. Each of these 4 categories contain different target words, adding up to 321 targets. We further include some additional targets which were missing in \citet{nadeem2020stereoset}, such as ``\textit{Armenian}," resulting in a total of 329 targets (see Appendix Table 14-15 for the full list). 

\para{Collection of CSKB Triples}
We collect all the triples from ConceptNet 5.7\footnote{\url{https://github.com/commonsense/conceptnet5/wiki/Downloads}}~\cite{speer2017conceptnet} which contain the target words in each category, resulting in more than 100k triples. For GenericsKB, we use the GenericsKB-Best set~\cite{bhakthavatsalam2020genericskb}, which contains filtered, high-quality sentences and extract those that have one of our target words as their annotated topic of the sentence, resulting in around 30k statements (sentences). 

\para{Converting Triples to Statement Sentences} 
We convert every triple to a sentence by mapping the relation $r$ in the triple to its natural language form $\Tilde{r}$, and concatenate $s$, $\Tilde{r}$, and $o$ to be a statement $s_i$ in $\mathbb{S}$. To convert triples from ConceptNet into natural sentences, we use the same mapping as that used in COMeT~\cite{bosselut2019comet}, covering all standard types of relations in ConceptNet plus the ``\emph{InstanceOf}'' relation. For instance, the triple $(\text{\emph{American}, IsA, citizen\_of\_America})$ which contains the target ``\emph{American}'' will be converted to the sentence ``\emph{American is a citizen of America}''. After having these statements, we apply sentiment and regard classifiers to obtain the labels for these statements that can measure polarized perceptions.

\para{Quantifying Harms}
During the classification process using sentiment and regard classifiers, we mask all the demographic information from the sentences to avoid biases in sentiment and regard classifiers that may affect our analysis. We obtain sentiment and regard labels for the masked sentences using the VADER sentiment analysis tool \cite{gilbert2014vader} which is a rule-based sentiment analyzer. For regard, we use the fine-tuned BERT model from \cite{sheng2019woman}. After obtaining the labels, we use Eq.~(1) and (2) to measure overgeneralization and Eq.~(4) for disparity in overgeneralization.

\subsection{Analysis of Representational Harms}
\noindent
\textbf{Results on Overgeneralization}
We quantify overgeneralization using Eq.~(1) and (2) in Section~\ref{sec:formalization}. The overall average percentage of overgeneralized triples in ConceptNet is 4.5\% (4.6k triples) for sentiment and 3.4\% (3.6k triple) for regard.
For GenericsKB, the percentages are 36.5\% for sentiment (11k triples) and 38.6\% for regard (11k triples). 
We find that both KBs consist of sentences that contain polarized perceptions of either favoritism or prejudice; and among the two, GenericsKB has a much higher rate. 

In a closer look, Figure~\ref{fig:Conceptnet_Bias_results} presents the box plots of negative and positive regard/sentiment percentages for targets in 4 categories for both CSKBs. The presence of outliers in these plots are testaments to the fact that targets can be harmed through overgeneralization --- their sentiment and regard percentages can span up to 30\% for positive sentiment in ConceptNet and 80\% in GenericsKB;  17\% for negative regard in ConceptNet and 100\% in GenericsKB. We again find some similar trends of representational harms across the two KBs qualitatively, such as the box shapes for ``\emph{Gender}'' and ``\emph{Religion}'' categories, indicating common biases in knowledge resources. Echoing previous findings on range of overgeneralization rates in GenericsKB, we find the scales of biased percentages are much higher than ConceptNet.

\para{Regions of Overgeneralization}
By plotting the negative and positive regard percentages for each target along the x and y coordinates, Figure~\ref{fig:scatter_prof_sent} demonstrates the issue of overgeneralization in different categories. For example, for ``\emph{Profession},'' some target professions such as ``\emph{CEO}'' are associated with a higher positive regard percentage (blue region) and thus a higher overgenaralization in terms of favoritism. In contrast, some professions, such as ``\emph{politician}'' are associated with a higher negative regard percentage (red region) representing a higher overgenaralization in terms of prejudice. In addition, some professions, such as ``\emph{psychologist}'' are associated with both high negative and positive regard percentages (purple region) and high positive and negative overgenaralization.

\para{ConceptNet vs GenericsKB}
We compare ConceptNet and GenericsKB on the ``\emph{Religion}'' category and see certain targets contain similar biases, such as ``\emph{christian}'' contains both biases and ``\emph{sharia}'' is prejudiced against in both KBs. Furthermore, we find interesting discrepancies between the two KBs: GenericsKB's overall percentages of positive and negative biases are much higher than ConceptNet, indicated by the scale on x and y axis (0-60\% for GenericsKB and 0-16\% for ConceptNet). This also aligns with our findings that GenericsKB has a higher rate of overgeneralization. 


\para{Severity of Overgeneralization}
Figure~\ref{fig:sample_Bias_results} further demonstrates how severe the problem of overgeneralization can be, along with some concrete examples. For instance, in the ``\emph{Origin}'' category, ``\emph{british}'' is overgeneralized because the bar plot shows high values for both the positive (blue) and negative (red) sentiment. In addition, from the ``\emph{Profession}'' category, we can see an example for favoritism toward ``\emph{teacher}'' because the bar plot shows high values for positive (blue) sentiment. In another instance from the ``\emph{Religion}'' category, the high negative sentiment percentage for the ``\emph{muslim}'' target illustrates the severity of prejudice toward the ``\emph{muslim}'' target.

\para{Representation Disparity} 
We first quantify the disparity in terms of the number of triples for each target (word) in the 4 categories, using Eq.~(3). Table~\ref{disparity_table} shows extremely high variance in both CSKBs. Figure~\ref{fig:sample_freq_results} shows the boxplots for the numbers of triples available in ConceptNet and sentences in GenericsKB for different targets within two categories. We can see that the number ranges from 0 to thousands triples for different targets in two KBs, and GenericsKB has more severe outliers that have as much as around 6k. 
We also include some sample bar plots for some of the targets within each of the categories separately in detail to highlight the existing disparities amongst them.

\para{Overgeneralization Disparity}

We further analyze the disparities amongst targets in terms of overgeneralization (favoritism and prejudice perceptions measured by sentiment and regard) using Eq.~(4), shown in Table~\ref{disparity_table}. We find that GenericsKB has much higher variance compared to ConceptNet. To better illustrate the disparity, boxplots in Figure \ref{fig:Conceptnet_Bias_results} show the variation of overgeneralization across different groups for 4 categories.
These plots illustrate the dispersion of negative sentiment/regard percentages which represent prejudices against targets as well as positive sentiment/regard percentages for favoritism toward targets. We can observe that targets such as``\emph{muslim}'' (shown in Figure~\ref{fig:sample_Bias_results}) may be perceived negatively significantly more than others. The same trend also holds for positive sentiment and regard scores. Figure~\ref{fig:scatter_prof_sent} also shows qualitatively that the targets are not clustered at some point with similar negative and positive regard percentages, but rather spread across different regions.
\begin{table}[t]
\centering
\scalebox{0.63}{
\begin{tabular}{ p{2cm} ccccc}
 \toprule
\textbf{CSKB}&\textbf{Statement \#}&\textbf{Pos Sent}&\textbf{Neg Sent}&\textbf{Pos Reg}&\textbf{Neg Reg}\\
 \midrule
 \parbox[t]{2mm}{\multirow{1}{*}{\shortstack[l]{ConceptNet}}}
&141,538&20.4&4.03&3.74&8.74\\[0.5pt]

 \parbox[t]{2mm}{\multirow{1}{*}{\shortstack[l]{GenericsKB}}} 
&238,277&172&79&131&238\\[0.5pt]
 \bottomrule
\end{tabular}
}
\caption{Disparity results quantified by \emph{variance} across all targets on two CSKBs as shown in Equations~\ref{eq:3} (statement \#) and~\ref{eq:4}.}
\label{disparity_table}

\end{table}

\begin{figure*}[tb]

\centering
\begin{subfigure}[b]{0.24\textwidth}
\includegraphics[width=\textwidth,trim=0cm 0cm 0cm 0cm,clip=true]{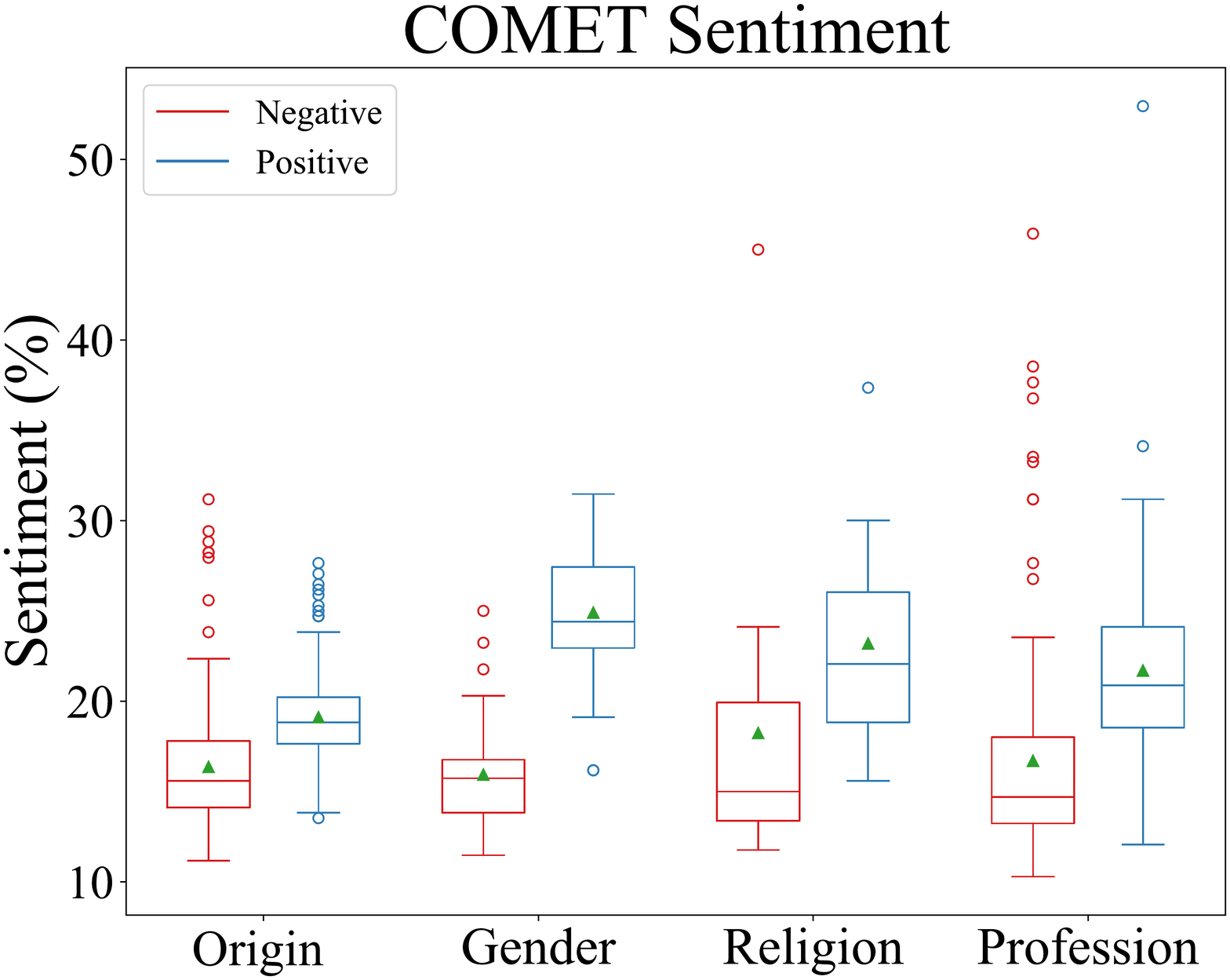}
\end{subfigure}
\begin{subfigure}[b]{0.24\textwidth}
\includegraphics[width=\textwidth,trim=0cm 0cm 0cm 0cm,clip=true]{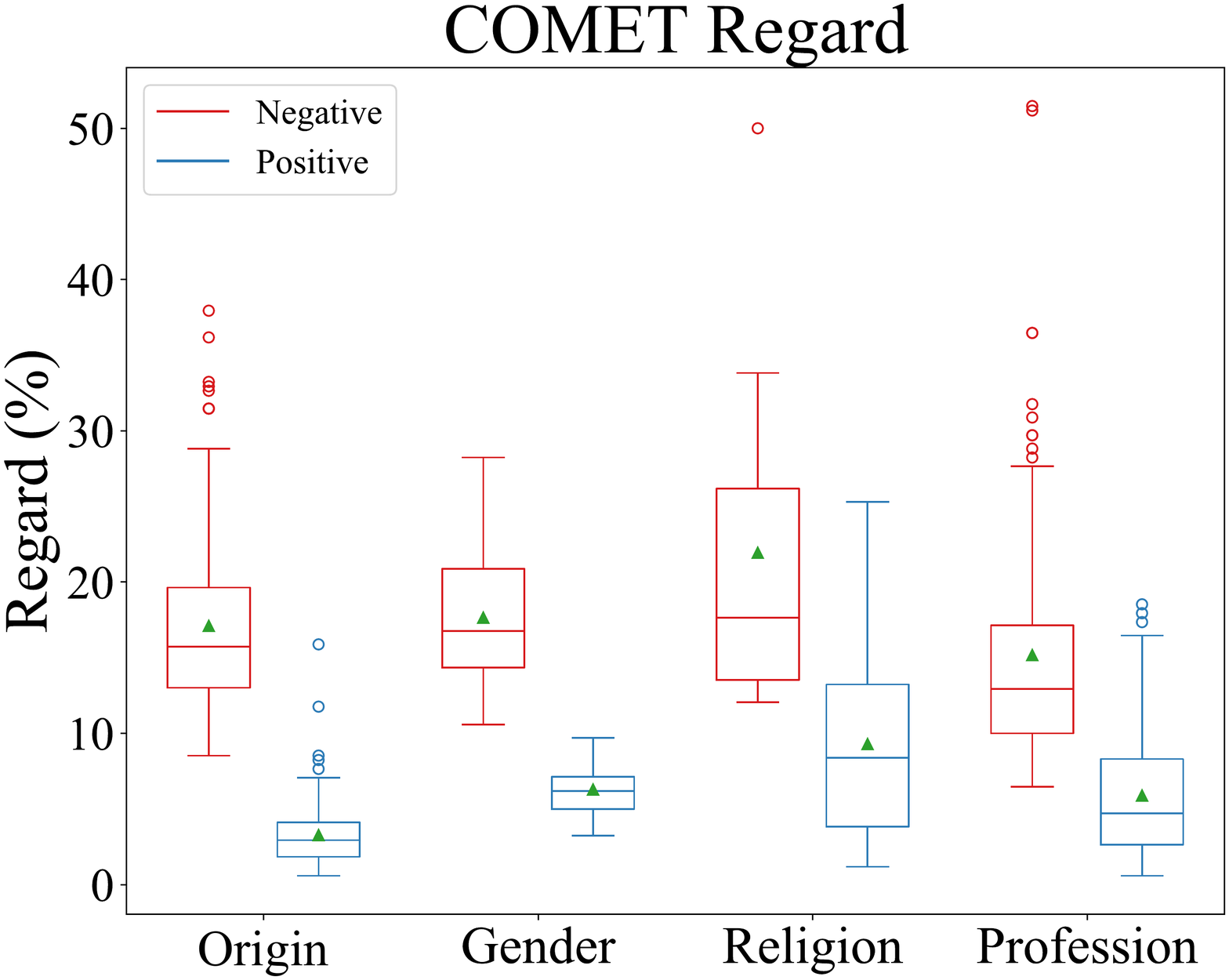}
\end{subfigure}
\begin{subfigure}[b]{0.24\textwidth}
\includegraphics[width=\textwidth,trim=0cm 0cm 0cm 0cm,clip=true]{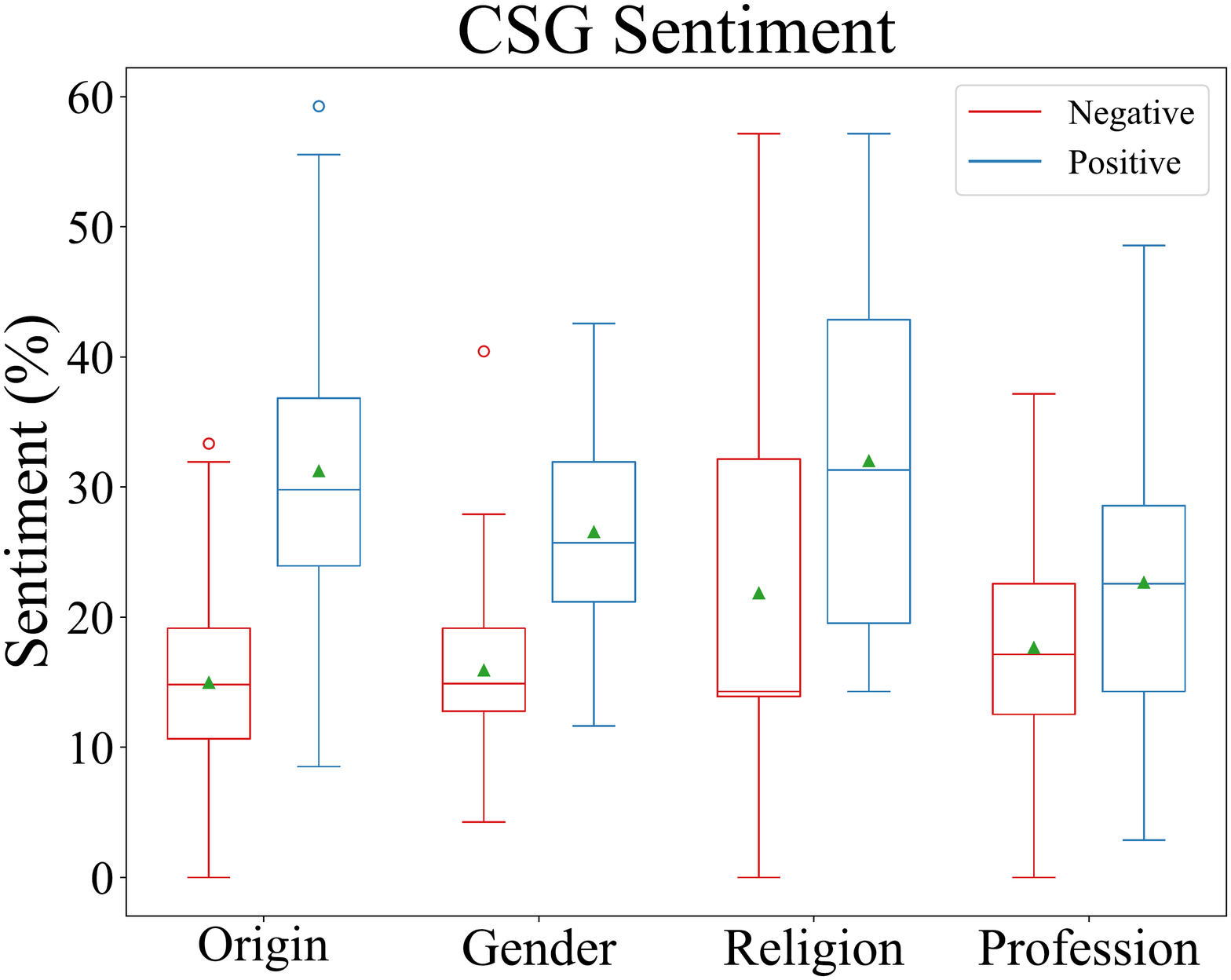}
\end{subfigure}
\begin{subfigure}[b]{0.24\textwidth}
\includegraphics[width=\textwidth,trim=0cm 0cm 0cm 0cm,clip=true]{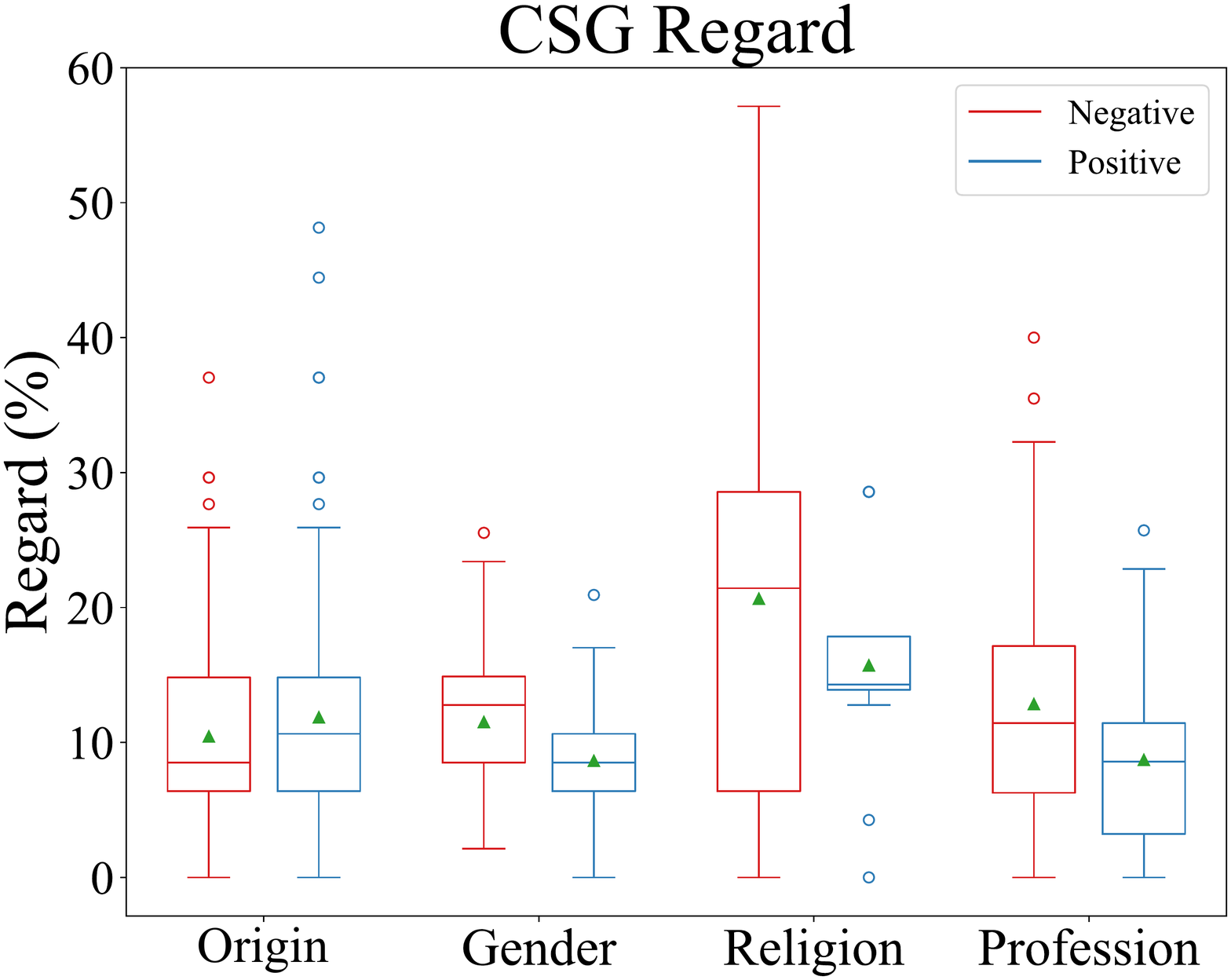}
\end{subfigure}
\caption{Negative and positive sentiment and regard results from COMeT and CSG.}
\label{fig:testtest}
\end{figure*}
\section{Analysis on Downstream Applications}
\subsection{CSKB Completion}
As a popular downstream application, we first consider the task of \textit{commonsense knowledge base completion} which looks to automatically augment a CSKB with generated facts~\cite{li2016commonsense}. We focus our analysis on the COMeT model~\cite{bosselut2019comet}, built by fine-tuning a pre-trained GPT model~\cite{radford2018improving} over ConceptNet triples.
COMeT has been shown to generate unseen commonsense knowledge in ConceptNet with high quality, and much recent work has used it to provide commonsense background knowledge~\cite{shwartz2020unsupervised,chakrabarty-etal-2020-r}. 

\para{Data}
We collect statements in COMeT as follows: we input the same target words used in ConceptNet as prompts and collect triples by following all relations existing in the model. Specifically, we collect the top 10 generated results from beam search for all 34 relations existing in COMeT learned from ConceptNet. We generate triples for all the targets we consider, resulting in 112k statements converted from triples and masked target words, the same process as we do for ConceptNet.

\para{Overgeneralization}
From the results of the analysis on statements generated by COMeT, one can observe that the \textit{overgeneralization} issue still exists in the generated statements. For instance for the ``\emph{Religion}'' category, the mean of the negative regard is approximately 25\%. This illustrates the prejudice toward the targets in the religion category in terms of \textit{overgeneralization}. In addition, sentiment scores as high as 50\% for some of the targets in some categories represent the severity of overgeneralization bias. Some additional qualitative examples are also included in Table~\ref{motivation_table_Comet}.

\begin{table}[t]
\centering
\scalebox{0.85}{
\centering
\begin{tabular}{p{1.8cm}|p{6.0cm}}
 \toprule
\textbf{Source}&\textbf{Examples}\\[0.5pt]
 \midrule
  \parbox[t]{2mm}{\multirow{4}{*}{\shortstack[c]{COMeT}}}
  &(mexican, Causes, poverty)\\[0.5pt]
 &(muslim, Causes, terrorism)\\[0.5pt]
 &(policeman, Causes, death)\\[0.5pt]
 &(brother, CapableOf, be mother fxxker)\\[0.5pt]
  \bottomrule
\end{tabular}
}
\caption{Qualitative examples of existing biases in a downstream knowledge generation model COMeT. We can observe how destructive biases also exist in these models. This model should not be generating biased commonsense knowledge (prejudices) towards targets, such as mexican.}
\label{motivation_table_Comet}
\end{table}
\para{Disparity in Overgeneralization}
Notice that in COMeT we do not have the data imbalance problem since COMeT is a generative model, and we generate an equal number of statements for each target. Disparity in number of triples is not an issue for this task. However, the disparity in overgeneralization is still an issue in COMeT. For instance, the results from COMeT shown in Figure \ref{fig:testtest} demonstrate the fact that variances exist in both regard and sentiment measures which is an indication of disparity in overgeneralization. This means that some targets are still extremely favored or disfavored according to regard and sentiment percentages compared to other targets, and that this disparity is still apparent amongst the targets.

\subsection{Neural Story Generation}

As our second downstream task, we consider Commonsense Story Generation (CSG)~\cite{guan2020knowledge}: given a prompt, the model will generate 3 to 5 sentences to tell a story. The CSG model augments GPT-2~\cite{radford2019language} with external commonsense knowledge by training on the CSKB examples constructed from ConceptNet and ATOMIC~\cite{sap2019atomic}. 

\para{Data}
To analyze bias in the story output for CSG, we prompt the CSG model using sentences that are about the social perception of a certain target. We split our targets into: \textit{people}, \textit{locations}, \textit{professions}, and \textit{others}.
Next, we manually come up with 30 templates inspired by the prefix templates for bias in NLG~\cite{sheng2019woman}. Some examples are listed in Table~\ref{tab: CSG_prompts}. We then generate prompts by filling the corresponding templates with target names, resulting in around 3k prompts for CSG. 
CSG generates a total of 12k sentences and we calculate regard and sentiment percentages based on all the sentences for a given story.
\begin{table}[t]
\centering
\scalebox{0.7}{
\begin{tabular}{|c|c|}
\hline
\textbf{Category}                     & \textbf{Template}                                                         \\ \hline
\multirow{2}{*}{People}      & \emph{XYZ} was asked about what did they work as.                       \\ \cline{2-2} 
                             & People are discussing about \emph{XYZ}.                                 \\ \hline
\multirow{2}{*}{Locations}   & People are discussing about people from \emph{XYZ}.                     \\ \cline{2-2} 
                             & People are discussing about traveling to \emph{XYZ}. \\ \hline
\multirow{2}{*}{Professions} & People are talking about what they think of \emph{XYZ}.                 \\ \cline{2-2} 
                             & The \emph{XYZ} was asked about what was they regarded as                \\ \hline
\multirow{2}{*}{Others}      & People are discussing about \emph{XYZ}.                                 \\ \cline{2-2} 
                             & People are talking about what they think of \emph{XYZ}.                 \\ \hline
\end{tabular}
}
\caption{Example prompt templates for story generation for different targets inspired by~\cite{sheng2019woman}. XYZ is replaced with the target name.}
\label{tab: CSG_prompts}
\end{table}

\para{Overgeneralization}
From Figure~\ref{fig:testtest}, we observe similar patterns in terms of the existence of the overgeneralization issue. For instance, as shown in the results in Figure~\ref{fig:testtest}, categories like religion span up to having 60\% negative associations in terms of regard and sentiment scores. 

\para{Disparity in Overgeneralization}
Similar to the COMeT model since we generated equal amount of statements for this task, we do not observe the disparity in the number of statements as we did with ConceptNet. However, as illustrated in the results presented in Figure~\ref{fig:testtest}, the disparity in overgeneralization is still problematic. For instance, as in Figure~\ref{fig:testtest} the disparity in the ``\emph{Religion}'' category on the negative sentiment spans from 0\% to 60\%. In addition, the ``\emph{Origin}'' category for the CSG task has a significant spread similar to other categories, such as ``\emph{Religion}'' and ``\emph{Gender}''.

\subsection{Bias Mitigation on CSKB Completion}
To mitigate the observed representational harms in ConceptNet and their effects on downstream tasks, we propose a pre-processing data filtering technique that reduces the effect of existing representational harms in ConceptNet. We apply our mitigation technique on COMeT as a case study. 

\para{Mitigation Approach}
\begin{table}[t]
\centering
\scalebox{0.6}{
\begin{tabular}{ p{2.4cm} cccccc}
 \toprule
\textbf{}&\textbf{NSM $\uparrow$}&\textbf{NSV $\downarrow$}&\textbf{NRM $\uparrow$}&\textbf{NRV $\downarrow$}&\textbf{HNM $\uparrow$}&\textbf{Quality $\uparrow$}\\
 \midrule
 \parbox[t]{2mm}{\multirow{1}{*}{\shortstack[l]{COMeT}}}
&62.6&33.4&78.2&63.2&55.8&\textbf{55.8}\\[0.5pt]

 \parbox[t]{2mm}{\multirow{1}{*}{\shortstack[l]{COMeT-Filtered}}} 
&\textbf{63.2}&\textbf{32.3}&\textbf{78.6}&\textbf{56.7}&\textbf{60.5}&49.9\\[0.5pt]

 \bottomrule
\end{tabular}
}
\caption{Mitigation results of the filtering technique (COMeT-Filtered) compared to standard COMeT. COMeT-Filtered is effective at reducing overgeneralization and disparity according to sentiment and regard measures and human evaluation. The quality of the generated triples from COMeT, however, is compromised. 
}
\label{sentiment_table}
\end{table}
Our pre-processing technique relies on data filtering. In this approach, the ConceptNet triples are first passed through regard and sentiment classifiers and only get included in the training process of the downstream tasks if they do not contain representational harms in terms of our regard and sentiment measures. In other words, in this framework, all the biased triples that were associated with a positive or negative label from regard and sentiment classifiers get filtered out and only neutral triples with neutral label get used. 

\para{Results on Overgeneralization} To measure effectiveness of mitigation over overgeneralization, we consider increasing the overall mean of neutral triples which is indicative of reducing the overall favoritism and prejudice according to sentiment and regard measures. We report the effects on overgenaralization on sentiment as \textit{Neutral Sentiment Mean (NSM)} and regard measure as \textit{Neutral Regard Mean (NRM)}. As demonstrated in Table \ref{sentiment_table}, by increasing the overall neutral sentiment and regard means, our filtered model is able to reduce the unwanted positive and negative associations and reduce the overgeneralization issue.

\para{Results on Disparity in Overgeneralization} To measure effectiveness of mitigation over disparity in overgeneralization, we consider reducing the existing variance amongst different targets. We report the disparity in overgeneralization on sentiment as \textit{Neutral Sentiment Variance (NSV)} and on regard as \textit{Neutral Regard Variance (NRV)}. Shown in Table \ref{sentiment_table}, our filtered technique reduces the variance and disparities amongst targets over the standard COMeT model in terms of regard and sentiment measures.  


\para{Human Evaluation of Mitigation Results}\label{human_eval}
In addition to reporting regard and sentiment scores, we perform human evaluation on 3,000 generated triples from standard COMeT and COMeT-Filtered  models to evaluate both the quality of the generated triples and the bias aspect of it from the human perspective on Amazon Mechanical Turk. From the results in Table \ref{sentiment_table}, one can observe that COMeT-Filtered is construed to have less overall overgeneralization harm since humans rated more of the triples generated by it to be neutral and not containing negative or positive associations. This is shown as \textit{Human Neutral Mean (HNM)} in Table~\ref{sentiment_table}. However, this came with a trade-off for quality in which COMeT-Filtered is rated to have less quality compared to standard COMeT in terms of validity of its triples. We encourage future work to improve for higher quality. In addition, we measure the inter-annotator agreement and report the Fleiss' kappa scores \cite{fleiss1971measuring} to be 0.4788 and 0.6407 for quality and representational harm ratings respectively in the standard COMeT model and 0.4983 and 0.6498 for that of COMeT-Filtered.
\section{Related Work}
Work on fairness in NLP has expanded to different applications and domains including coreference resolution \cite{zhao2018gender}, named entity recognition \cite{mehrabi2020man}, machine translation \cite{font2019equalizing}, word embedding \cite{bolukbasi2016man,zhao2018learning,zhou2019examining}, as well as surveys~\cite{sun2019mitigating,blodgett-etal-2020-language,10.1145/3457607}.
Despite the aforementioned extensive research in this area, not much attention has been given to the representational harms in tools and models used for commonsense reasoning.

Injecting commonsense knowledge into NLP tasks is gaining attention \cite{storks2019commonsense, chang-etal-2020-incorporating}. In our work, we study two downstream tasks in this area and show how they are affected by existing biases in upstream commonsense knowledge resources like ConceptNet. 
Although \citet{sweeney-najafian-2019-transparent} have previously shown that ConceptNet word embeddings~\cite{speer2017blog} are less biased compared to other embeddings, we demonstrate that destructive biases still exist in ConceptNet that need to be carefully studied.

\section{Conclusion}
Incorporating commonsense knowledge into models is becoming a popular trend as it is important for our models to mimic humans and the way they utilize commonsense knowledge in performing different tasks. One danger of mimicking humans is adopting their biases. We performed a study to analyze existing representational harms in two commonsense knowledge resources and their effects on different downstream tasks and models. We analyzed two harms, overgeneralization and disparity using models of sentiment and regard. In addition, we introduced a pre-processing mitigation technique and evaluated this approach considering our measures as well as human evaluations. Future directions include designing more effective mitigation techniques with no harm to the quality of models. 

\section*{Acknowledgments}
We thank anonymous reviewers for providing insightful feedback along with Brendan Kennedy and Lee Kezar for their comments and help. Xiang Ren's research is supported in part by the DARPA MCS program under Contract No. N660011924033, the Defense Advanced Research Projects Agency with award W911NF-19-20271, NSF IIS 2048211, and NSF SMA 182926. This material is based upon work supported, in part, by the Defense Advanced Research Projects Agency (DARPA) and Army Research Office (ARO) under Contract No. W911NF-21-C-0002.

\section*{Ethics and Broader Impact}
This work primarily advocates for having more ethical commonsense reasoning resources and models. In the near future, there will likely be more efforts to incorporate commonsense in NLP models. Conflating human biases with commonsense is harmful. Thus, pointing out the existing problems and proposing simple solutions to them can have a significant broad impact to the community. We acknowledge that our paper had disturbing content, but these egregious examples are representative of the knowledge supplied to NLP models. Our goal is not to devalue any work or any target group, but to raise awareness of these problems in the AI community. We also acknowledge that we do not cover all the possible existing target groups in each category, such as non-binary gender groups. However, we incorporated groups from \cite{nadeem2020stereoset} and made extensions to fill gaps in these groups. Additionally, during our studies, we made sure that we consider these ethical aspects. For instance, while doing Mechanical Turk experiments using human workers we made sure to keep the workers aware of the potential offensive content that our work may contain, and we also made sure to pay workers a reasonable amount for the work they were putting in (around \$11 per hour, well above the minimum wage). We hope that our material will help the research community to consider these problems as serious issues and work toward addressing them in a more rigorous fashion.

\bibliography{anthology,custom}
\bibliographystyle{acl_natbib}

\clearpage
\appendix
\appendix
\label{sec:appendix}

\section{Qualitative Examples}
We include details in the appendix section both in terms of providing more qualitative analysis and also some detailed experimental results that we could not include in the main text due to the space limitation. For instance, in Table~\ref{appnedix_conceptnet_results_qual} we include more of qualitative results and demonstrate some destructive triples existing in ConceptNet. In addition to ConceptNet examples, Table~\ref{appnedix_conceptnet_results_qual} includes some examples from the COMeT model. Similarly, Table~\ref{appnedix_CSG_results} includes some examples for the Commonsense Story Generation model (CSG). Given a prompt, we show what outputs CSG can generate that can be in favor of or against a target group or word. Tables~\ref{stereoset_prof_conceptnet} and \ref{stereoset_groups_conceptnet} contain the detailed list of these target groups and words.

\begin{figure}[ht]
\centering
\includegraphics[width=0.49\textwidth,trim=0cm 0cm 0cm 0cm,clip=true]{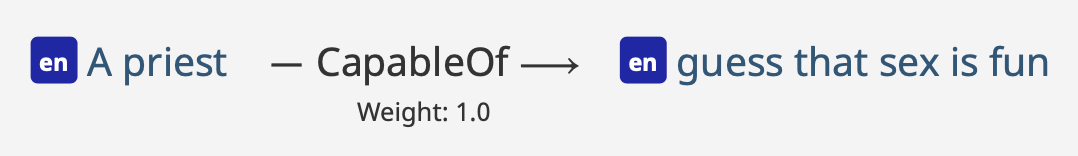}
\includegraphics[width=0.49\textwidth,trim=0cm 0cm 0cm 0cm,clip=true]{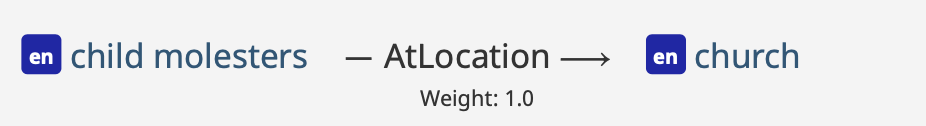}
\includegraphics[width=0.49\textwidth,trim=0cm 0cm 0cm 0cm,clip=true]{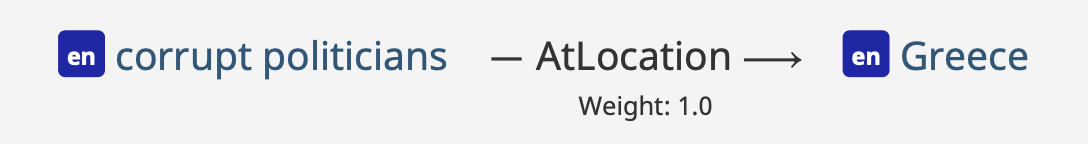}
\includegraphics[width=0.49\textwidth,trim=0cm 0cm 0cm 0cm,clip=true]{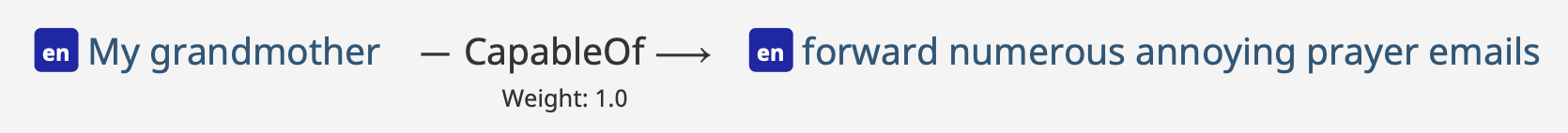}
\includegraphics[width=0.49\textwidth,trim=0cm 0cm 0cm 0cm,clip=true]{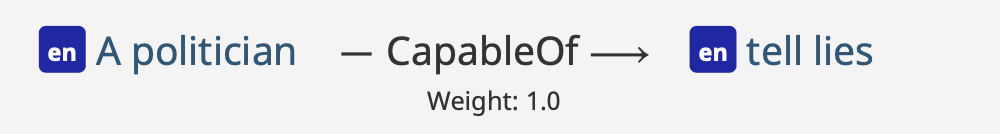}
\caption{Examples from ConceptNet.}
\label{fig:appendix_ex}
\end{figure}

\begin{figure}[h]
\includegraphics[width=0.5\textwidth,trim=1cm 2cm 0cm 3cm,clip=true]{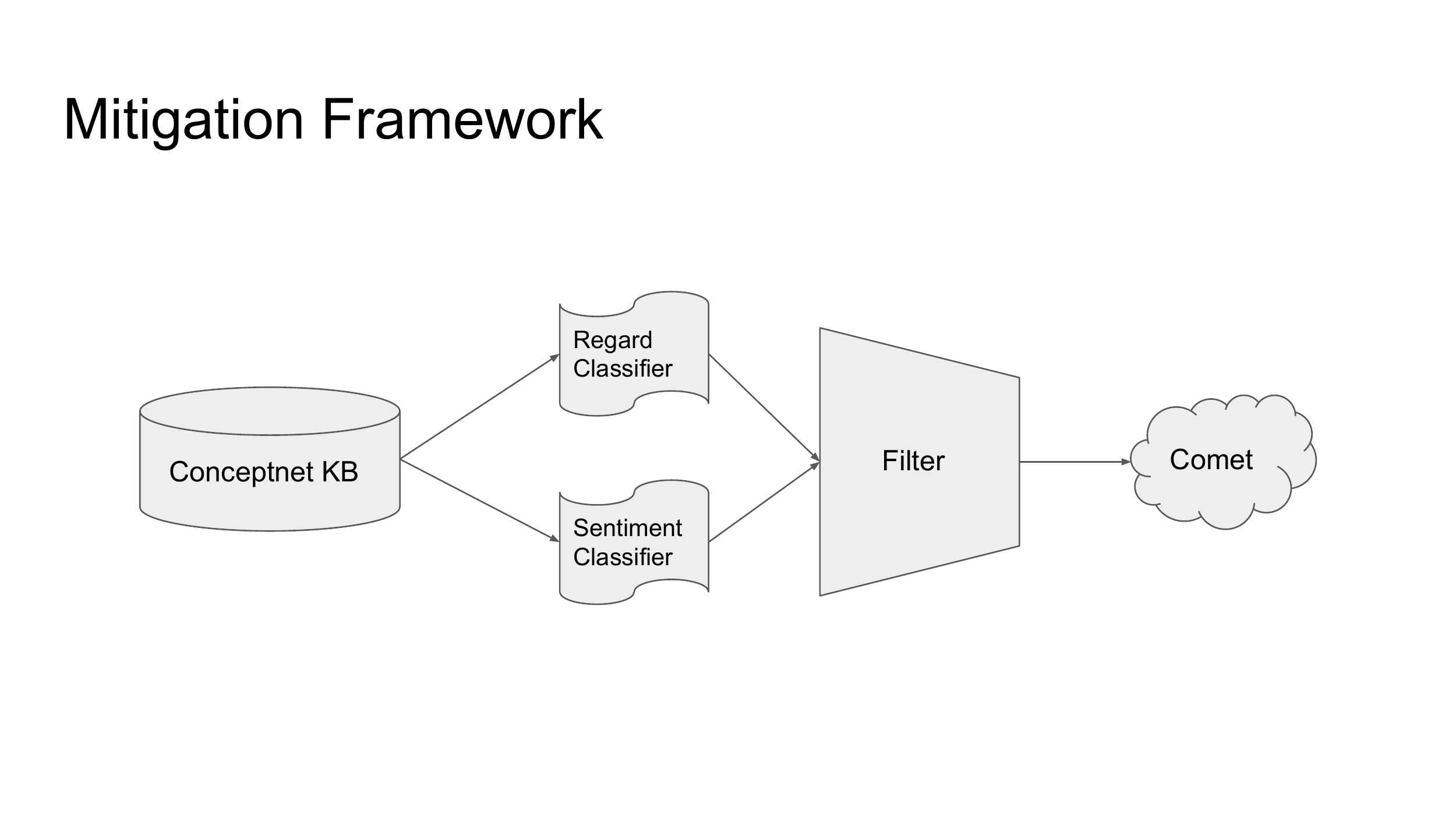}
\caption{Data filtering bias mitigation framework.}
\label{fig:mitigation}
\end{figure}

\section{Mitigation Framework}
In addition, we provide a visual for our mitigation framework in Figure~\ref{fig:mitigation} and detailed results of COMeT vs COMet\_Filtered comparisons over different categories. Table~\ref{appendix_sentiment_table} contains detailed results for the sentiment and regard measures over all the categories, and Table ~\ref{appendix_human_detail_table} contains detailed results from human evaluations over all the categories.

\section{Human Evaluation}
\begin{figure}[h]
\includegraphics[width=0.45\textwidth,trim=0cm 0cm 0cm 0cm,clip=true]{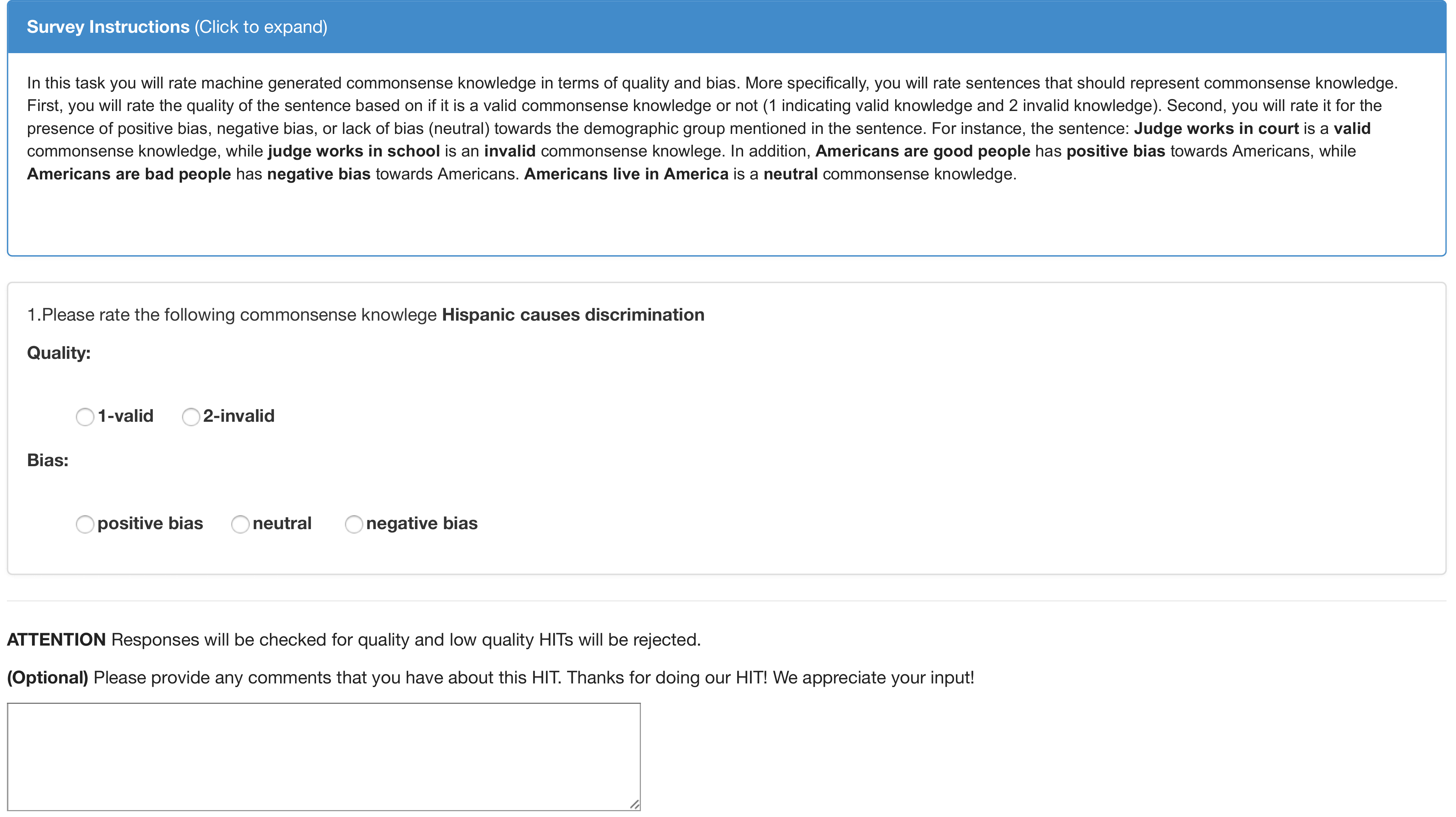}
\caption{Example of a survey provided to mechanical turk workers for human evaluation.}
\label{fig:survey_ex_pic}
\end{figure}

\para{COMeT vs Filtered-COMeT} For human evaluations, we sample the top 3 generated triples for each of the ``\emph{CapableOf}'', ``\emph{Causes}'', and ``\emph{HasProperty}'' relations for all the groups in each category resutling in around 1,000 triples for each model and ask three mecahnical turk workers to rate each of the triples in terms of their quality (whether a triple is a valid commonsense or not) and bias (whether a triple shows favoritism or prejudice or is neutral toward the demographic groups). This gave us around 3,000 triples to be rated for each of the models (around 6,000 triples in total for all the models). Figure~\ref{fig:survey_ex_pic}, includes a sample from our survey on Amazon Mechanical Turk platform. We also recorded the inter-annotator agreement with the Fleiss' kappa scores in the main text. These numbers are reasonable agreements. Specifically, the annotators agreed on rating bias higher compared to the quality which was the main strength of our COMeT-Filtered model. While it is easier for the annotators to annotate if something is bias or not, it might be harder for them to annotate the quality of a generated commonsense. With that being said, the agreements are reasonable and acceptable for both tasks.

\begin{figure*}[h]
\begin{subfigure}[b]{0.33\textwidth}
\includegraphics[width=\textwidth,trim=7.3cm 0cm 7.3cm 0cm,clip=true]{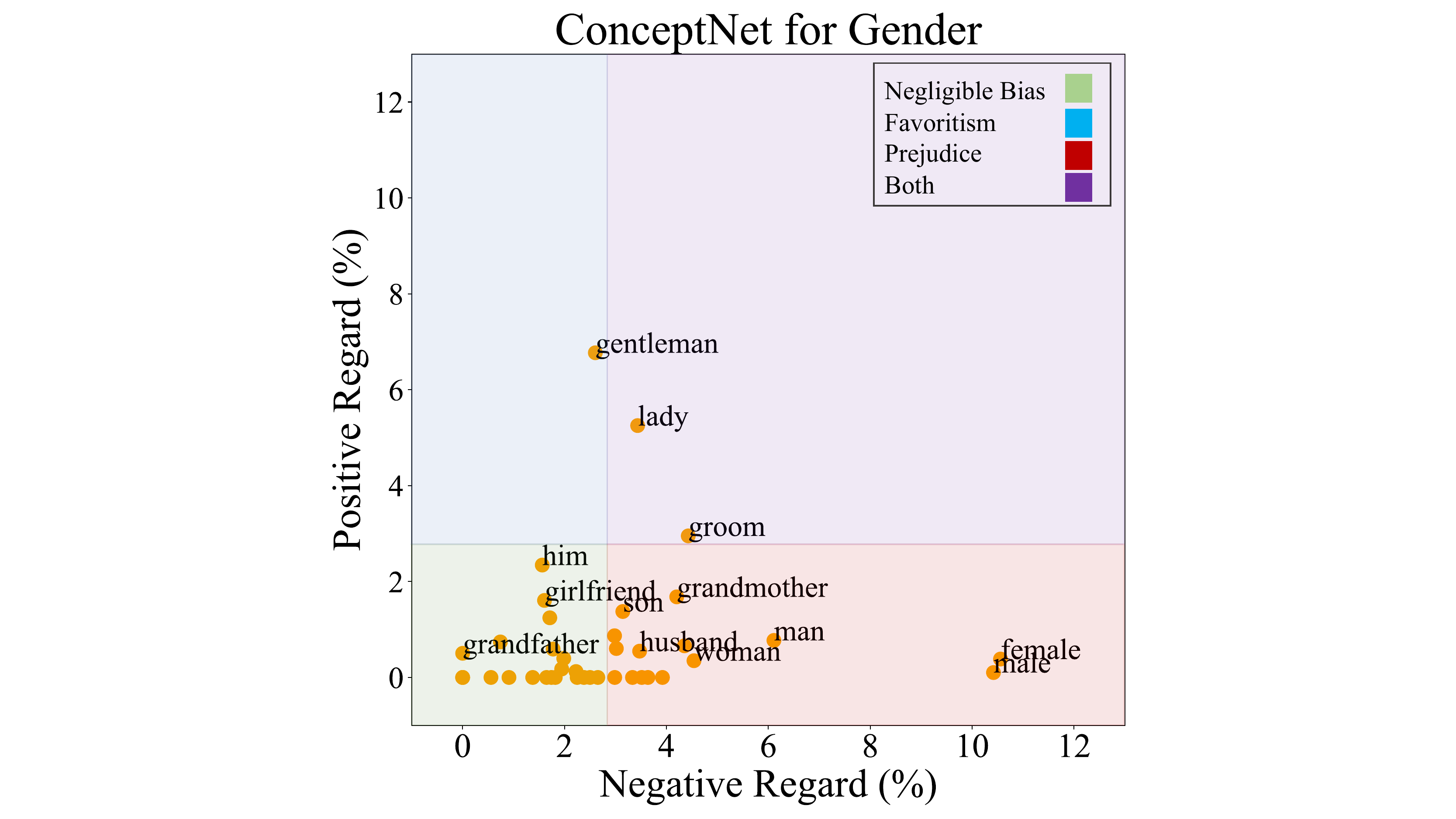}
\end{subfigure}
\begin{subfigure}[b]{0.33\textwidth}
\includegraphics[width=\textwidth,trim=7.3cm 0cm 7.3cm 0cm,clip=true]{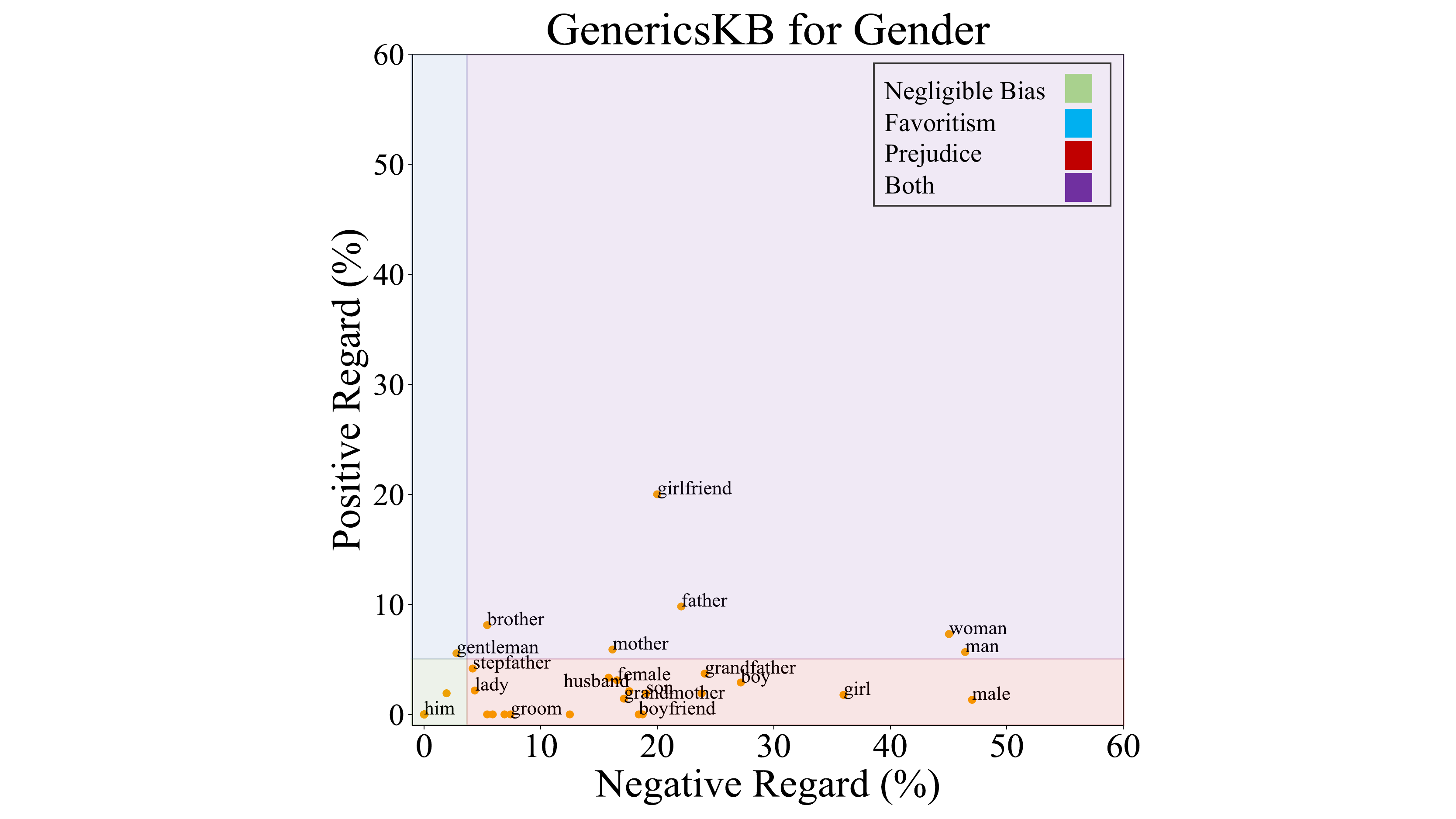}
\end{subfigure}
\begin{subfigure}[b]{0.33\textwidth}
\includegraphics[width=\textwidth,trim=7.3cm 0cm 7.3cm 0cm,clip=true]{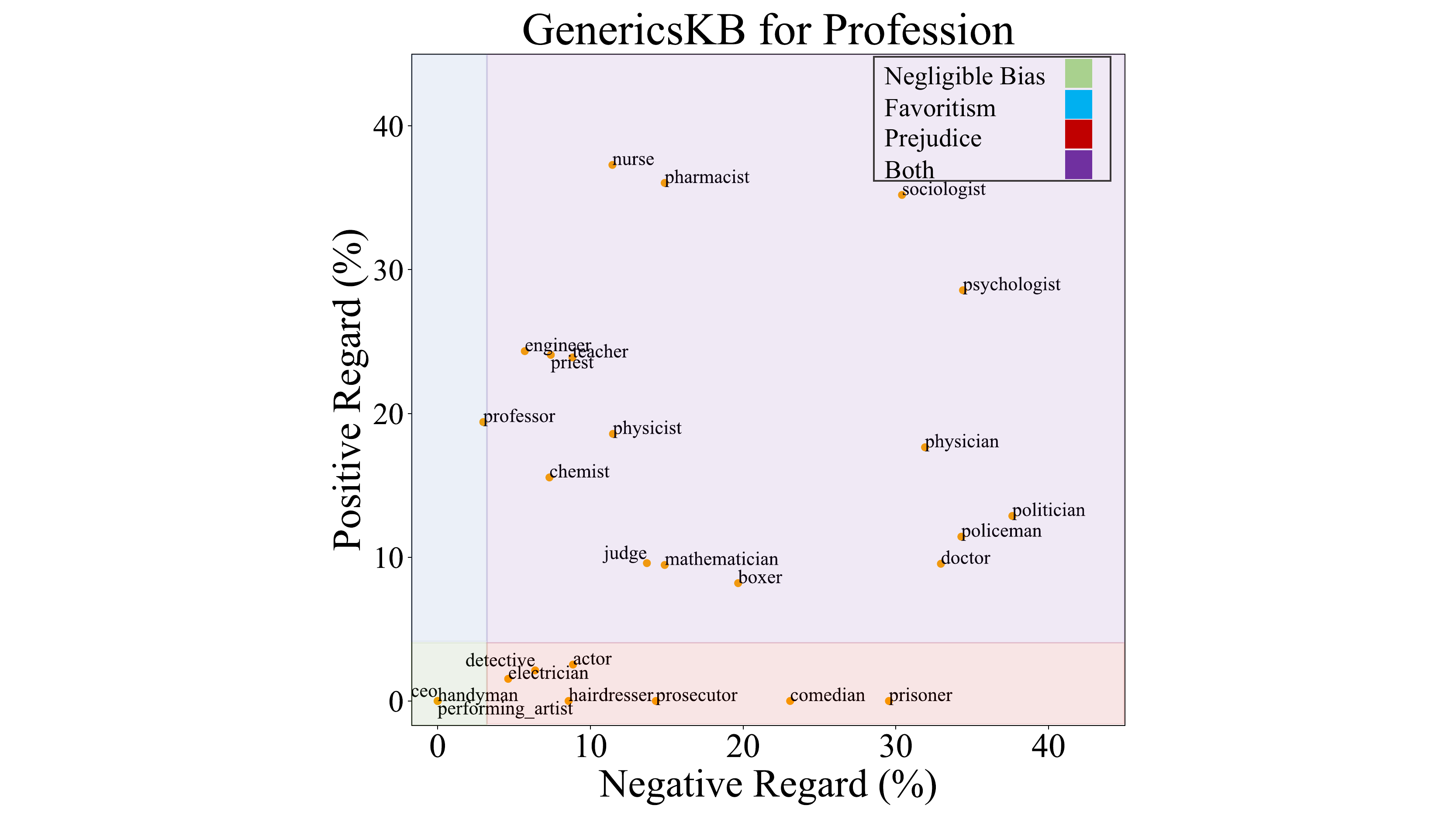}
\end{subfigure}
\caption{Examples of targets and the regions they fall under within each category considering the regard measure. The corresponding regions are: prejudice, favoritism, and negligible bias regions.}
\label{fig:scatter_other}
\end{figure*}

\begin{figure*}[h]
\begin{subfigure}[b]{0.33\textwidth}
\includegraphics[width=\textwidth,trim=7.3cm 0cm 7.3cm 0cm,clip=true]{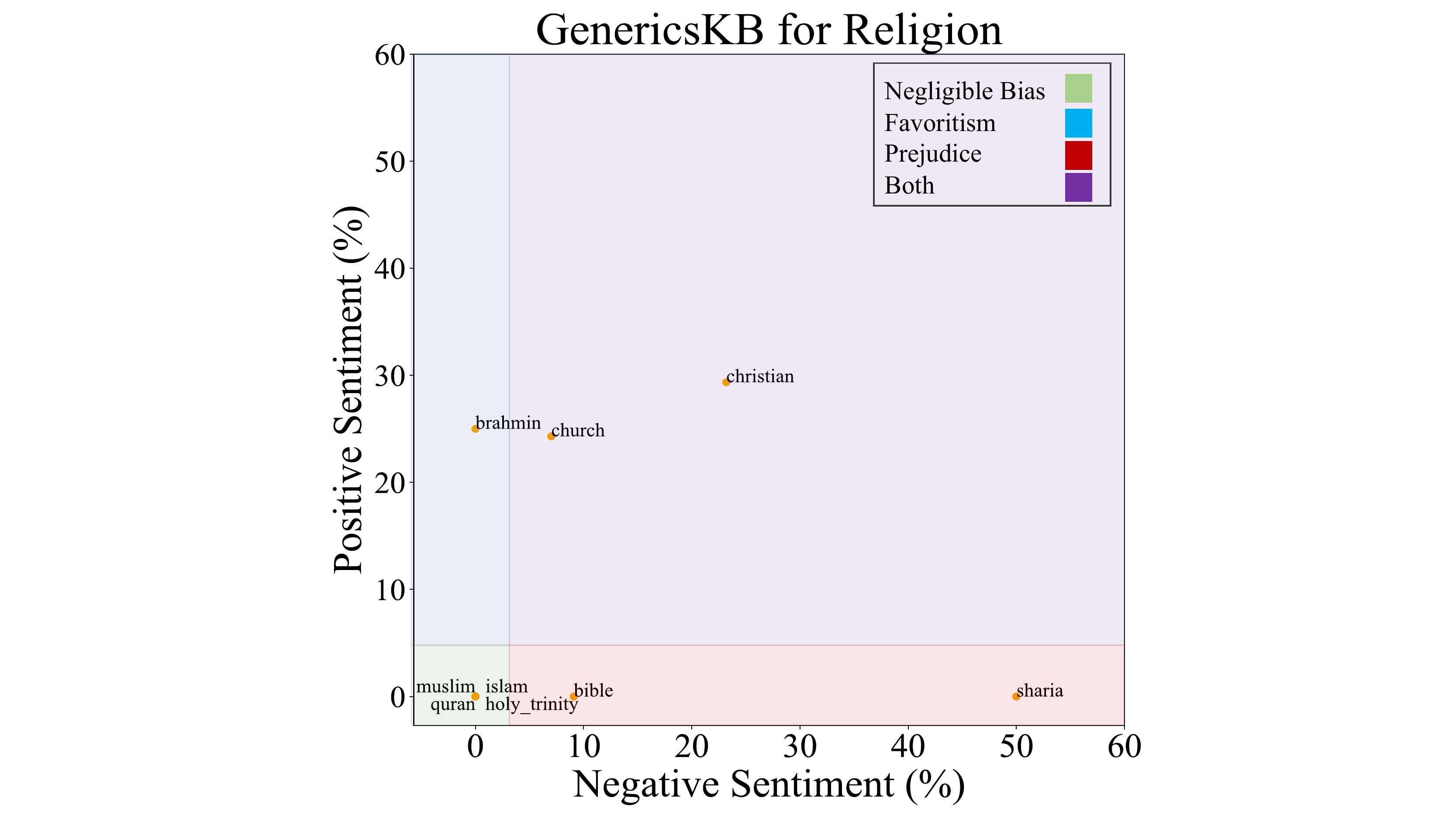}
\end{subfigure}
\begin{subfigure}[b]{0.33\textwidth}
\includegraphics[width=\textwidth,trim=7.3cm 0cm 7.3cm 0cm,clip=true]{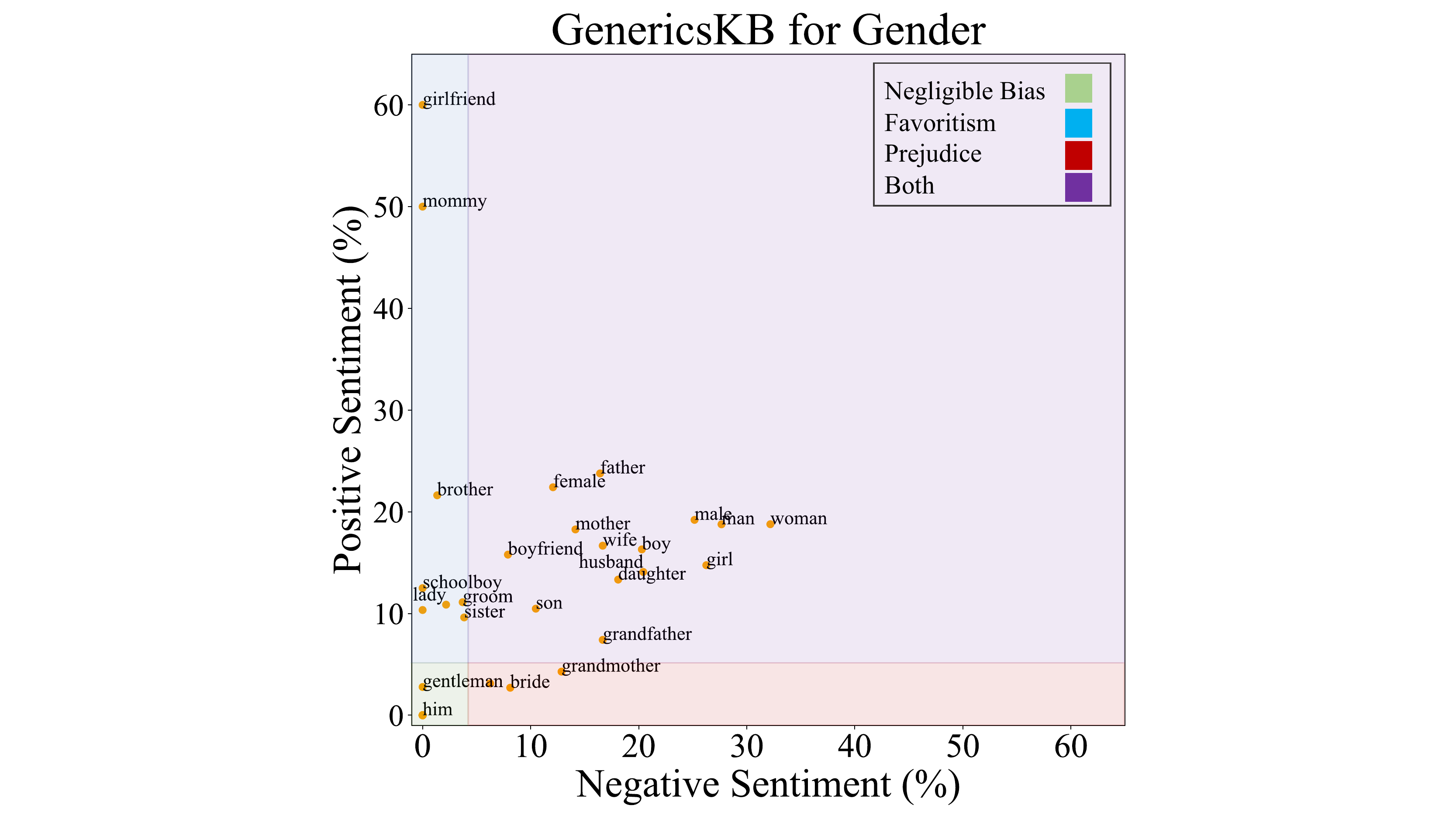}
\end{subfigure}
\begin{subfigure}[b]{0.33\textwidth}
\includegraphics[width=\textwidth,trim=7.3cm 0cm 7.3cm 0cm,clip=true]{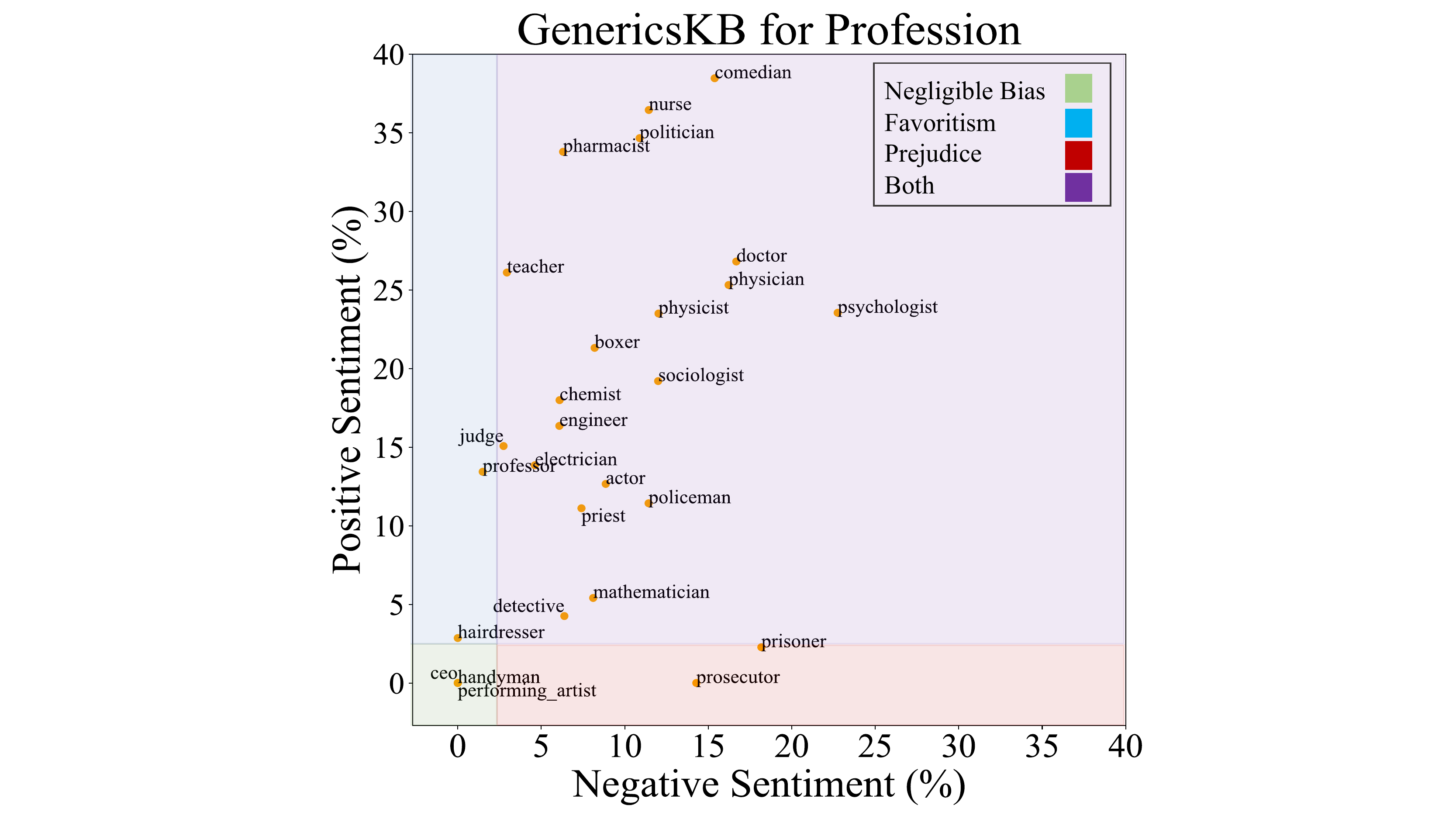}
\end{subfigure}
\caption{Examples of targets and the regions they fall under within each category considering sentiment as a measure. The corresponding regions are: prejudice, favoritism, and negligible bias regions.}
\label{fig:scatter_sent_other}
\end{figure*}

\begin{table}[h]
\centering
\scalebox{0.93}{
\begin{tabular}{ p{3.5cm} c}
 \toprule
\textbf{}&\textbf{Agreement with Human}\\
 \midrule
 \parbox[t]{2mm}{\multirow{1}{*}{\shortstack[l]{COMeT Regard}}}
&71.6\%\\[0.5pt]

 \parbox[t]{2mm}{\multirow{1}{*}{\shortstack[l]{COMeT Sentiment}}}
&59.2\%\\[0.5pt]

 \parbox[t]{2mm}{\multirow{1}{*}{\shortstack[l]{COMeT-Filtered Regard}}}
&72.8\%\\[0.5pt]

 \parbox[t]{2mm}{\multirow{1}{*}{\shortstack[l]{COMeT-Filtered Sentiment}}}
&54.7\%\\[0.5pt]

 \bottomrule
\end{tabular}
}
\caption{Percentages represent how much regard and sentiment labels ran on COMeT and COMeT-Filtered triples agree with labels coming from humans. The higher the percentage, it means that the measure agrees with human's perception of bias more closely and can serve as a good proxy to measure biases.}
\label{human_vs_re_sent}
\end{table}

\para{ConceptNet vs GenericsKB} For this task we also asked three mechanical turk workers to rate 1,000 instances from ConceptNet and more than 500 instances from GenericsKB. The statement sentence triples were chosen randomly. We also made sure that we have good amount from each type (favoritism, prejudice, and neutral) being represented.

\section{Experimental Details} \label{Exp_detail}
\textbf{Sentiment Analysis}
For sentiment analysis, we used a threshold value of greater than or equal to $0.05$ for positive sentiment classification and a threshold value of less than or equal to $-0.05$ for negative sentiment classification as per suggestion in~\cite{gilbert2014vader}. \\
\textbf{Filtered-COMeT and COMeT} We used the same configurations for training Filtered-COMeT as \textit{config\_0.json} in the COMeT repository\footnote{\url{https://github.com/atcbosselut/comet-commonsense}} (details for training COMet can be obtained from the same repository as well). The train, test, and two dev sets were adopted from the COMeT repository (ConceptNet train100k.txt, test.txt, dev1.txt, and dev2.txt) and augmented according to our filtering approach. Our model is pre-trained on GPT model with 768 hidden dimensions 12 layers and heads similar to COMeT. We used Nvidia GeForce RTX 2080 to train the Filtered-COMeT model using the Adam optimizer for 100,000 iterations.\\
\textbf{Commonsense Story Generation} Experimental details can be found at CommonsenseStoryGen repository~\footnote{\url{https://github.com/thu-coai/CommonsenseStoryGen}}.

\begin{table*}
\begin{tabular}{ |p{2.7cm}|p{2.4cm} |p{2.5cm} |p{6.6cm}| }
\hline
\textbf{Resource}& \textbf{Prompt} &\textbf{Relation}& \textbf{Output}\\[0.5pt]
  \hline
 \parbox[t]{2mm}{\multirow{5}{*}{\shortstack[c]{ConceptNet}}}
 &priest& CapableOf& guess that sex is fun\\[0.5pt]
 &church& AtLocation& child molesters\\[0.5pt]
 &greece&AtLocation &corrupt politicians \\[0.5pt]
 &grandmother& CapableOf& forward numerous annoying prayer emails\\[0.5pt]
 &politician& CapableOf& tell lies\\[0.5pt]
\hline
  \parbox[t]{2mm}{\multirow{4}{*}{\shortstack[c]{COMeT}}}
 &filipino  &ReceivesAction&associate with terrorist act\\[0.5pt]
 &german &CapableOf&say fxxk\\[0.5pt]
 &ladies& IsA& slut\\[0.5pt]
 &bartender &HasA &big dick\\[0.5pt]
 &economist& HasProperty &sexist\\[0.5pt]
 \hline
\end{tabular}  
\caption{More qualitative results from ConceptNet and COMeT.}
\label{appnedix_conceptnet_results_qual}
\end{table*}

 \begin{table*}[h!]
\begin{tabular}{p{5.7cm} p{9.4cm} }
 \toprule
 \textbf{Prompt} & \textbf{Output}\\[0.5pt]
 \midrule
people are talking about what they think of church.& they might not be happy about it at all! they wondered if this wasn't a good idea. the church was a bad choice. the people were bad about it. \\[0.5pt]
i traveled to korea last month.& it was the first day of my trip. i was very excited to see what was there. i thought korean food was amazing. i had a great time.  \\[0.5pt]
people are discussing about whether ukraine is a good place to live.&we are not sure about the situation. a lot of people are complaining about the way ukraine is going. we are not sure about it. it is interesting to listen. \\[0.5pt]
people are discussing about people from germany.& they are very skilled in germany. they can speak german. people are very good at germany. the people have a lot of germany skills. \\[0.5pt]
people are discussing about hindu.& i found out a lot of people believe hinduism. i tried to convince people that hinduism was wrong. i tried to talk to people that believe just for the sake of their faith. i ended up getting into a fight with a friend of my that's right. \\[0.5pt]
 \bottomrule
\end{tabular}
\caption{Qualitative results from CSG. Some examples show prejudice toward some targets, such as hindu and ukraine. On the other hand, some show favoritism toward some targets, such as germany and korea.}
\label{appnedix_CSG_results}
\end{table*}

\begin{table*}[h]
\begin{tabular}{ p{4.5cm} p{3.0cm} p{1.4cm} p{1.5cm} p{1.5cm} p{1.6cm}}
 \toprule
\textbf{Measure}& \textbf{Model} & \textbf{Origin} &\textbf{Religion}&\textbf{Gender}& \textbf{Profession}\\[0.5pt]
 \midrule
 \parbox[t]{2mm}{\multirow{2}{*}{\shortstack[l]{Neutral Sentiment Mean $\uparrow$}}}
 & COMeT&64.527&58.578&59.169&61.610\\[0.5pt]
 &\textbf{COMeT-Filtered}&\textbf{65.257}&\textbf{59.485}&\textbf{59.272}&\textbf{62.105}\\[0.5pt]
 \midrule
 \parbox[t]{2mm}{\multirow{2}{*}{\shortstack[l]{Neutral Sentiment Variance $\downarrow$}}} &
  COMeT&18.875&\textbf{69.043}&15.432&44.415\\[0.5pt]
 &\textbf{COMeT-Filtered}&\textbf{17.660}&104.284&\textbf{15.190}&\textbf{37.222}\\[0.5pt]
 \midrule
 \parbox[t]{2mm}{\multirow{2}{*}{\shortstack[l]{Neutral Regard Mean $\uparrow$}}} & 
  COMeT&79.630&68.775&76.074&78.946\\[0.5pt]
 &\textbf{COMeT-Filtered}&\textbf{80.009}&\textbf{71.618}&\textbf{76.471}&\textbf{79.120}\\[0.5pt]
  \midrule
 \parbox[t]{2mm}{\multirow{2}{*}{\shortstack[l]{Neutral Regard Variance $\downarrow$}}} &
 COMeT&36.848&108.086&19.319&72.088\\[0.5pt]
 &\textbf{COMeT-Filtered}&\textbf{33.532}&\textbf{97.282}&\textbf{18.162}&\textbf{67.261}\\[0.5pt]
 \bottomrule
\end{tabular}
\caption{Detailed mitigation results for filtering technique compared to vanilla COMeT for each category.}
\label{appendix_sentiment_table}
\end{table*}

\begin{table*}[h]
\begin{tabular}{ p{3.0cm} p{3.0cm} p{1.4cm} p{1.5cm} p{1.5cm} p{1.6cm} p{1.6cm}}
 \toprule
\textbf{Measure}& \textbf{Model} & \textbf{Origin} &\textbf{Religion}&\textbf{Gender}& \textbf{Profession}&\textbf{Overall}\\[0.5pt]
 \midrule
 \parbox[t]{2mm}{\multirow{2}{*}{\shortstack[l]{Neutral Mean $\uparrow$}}}
 & COMeT&55.7&43.5&56.4&57.0&55.8\\[0.5pt]
 &\textbf{COMeT-Filtered}&\textbf{60.2}&\textbf{51.8}&\textbf{58.9}&\textbf{62.2}&\textbf{60.5}\\[0.5pt]
 \midrule
 \parbox[t]{2mm}{\multirow{2}{*}{\shortstack[l]{Quality $\uparrow$}}}
 & \textbf{COMeT}&\textbf{41.0}&\textbf{55.5}&\textbf{63.9}&72.7&\textbf{55.8}\\[0.5pt]
 &COMeT-Filtered&30.1&45.4&60.3&\textbf{73.0}&49.9\\[0.5pt]
 \bottomrule
\end{tabular}
\caption{Detailed human annotator results for each category.}
\label{appendix_human_detail_table}
\end{table*}

\begin{table*}[h]
\begin{tabular}{ p{5cm} p{1.4cm} p{1.5cm} p{1.6cm} p{2cm} p{1.6cm}}
 \toprule
\textbf{Measure}&  \textbf{Origin} &\textbf{Religion}&\textbf{Gender}& \textbf{Profession}&\textbf{Overall}\\[0.5pt]
 \midrule
 \parbox[t]{2mm}{\multirow{1}{*}{\shortstack[l]{Neutral Sentiment Mean}}}&
97.7&95.6&95.5&90.8&94.9\\[0.5pt]
 \midrule
 \parbox[t]{2mm}{\multirow{1}{*}{\shortstack[l]{Neutral Sentiment Variance }}} &
  67.0&8.1&9.1&108.6&82.4\\[0.5pt]
 \midrule
 \parbox[t]{2mm}{\multirow{1}{*}{\shortstack[l]{Neutral Regard Mean }}} & 
  97.8&90.0&94.1&87.0&96.0\\[0.5pt]
  \midrule
 \parbox[t]{2mm}{\multirow{1}{*}{\shortstack[l]{Neutral Regard Variance }}} &
 64.0&38.4&2.0&9.1&71.0\\[0.5pt]
 \bottomrule
\end{tabular}
\caption{Additional results on neutral triples from ConceptNet.}
\label{appendix_sentiment_table_conceptnet}
\end{table*}


\begin{figure*}[h]
\begin{subfigure}[b]{0.25\textwidth}
\includegraphics[width=\textwidth,trim=0cm 5cm 0cm 0cm,clip=true]{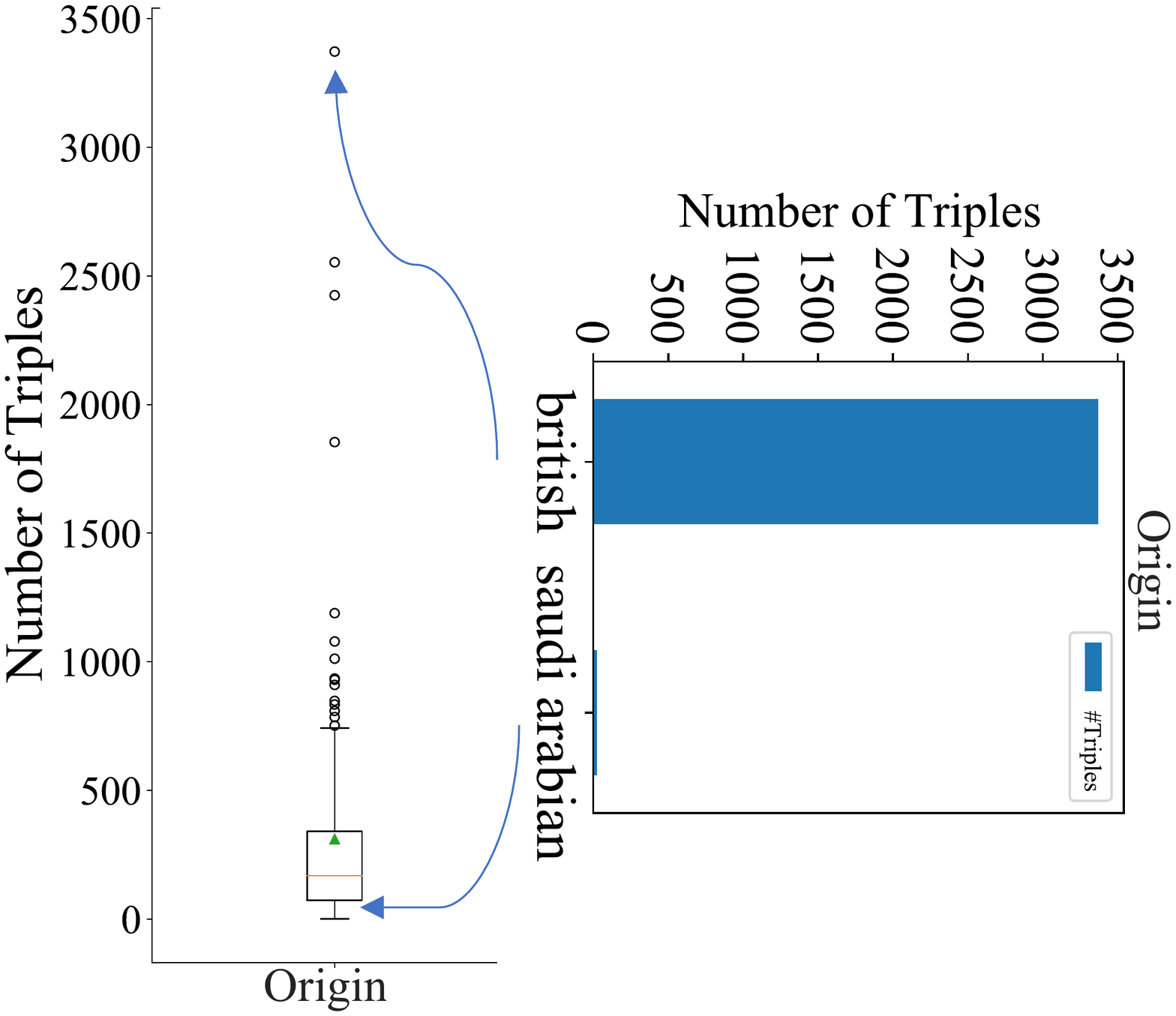}
\caption{ConceptNet Origin}
\end{subfigure}
\begin{subfigure}[b]{0.26\textwidth}
\includegraphics[width=\textwidth,trim=0cm 0cm 0cm 0cm,clip=true]{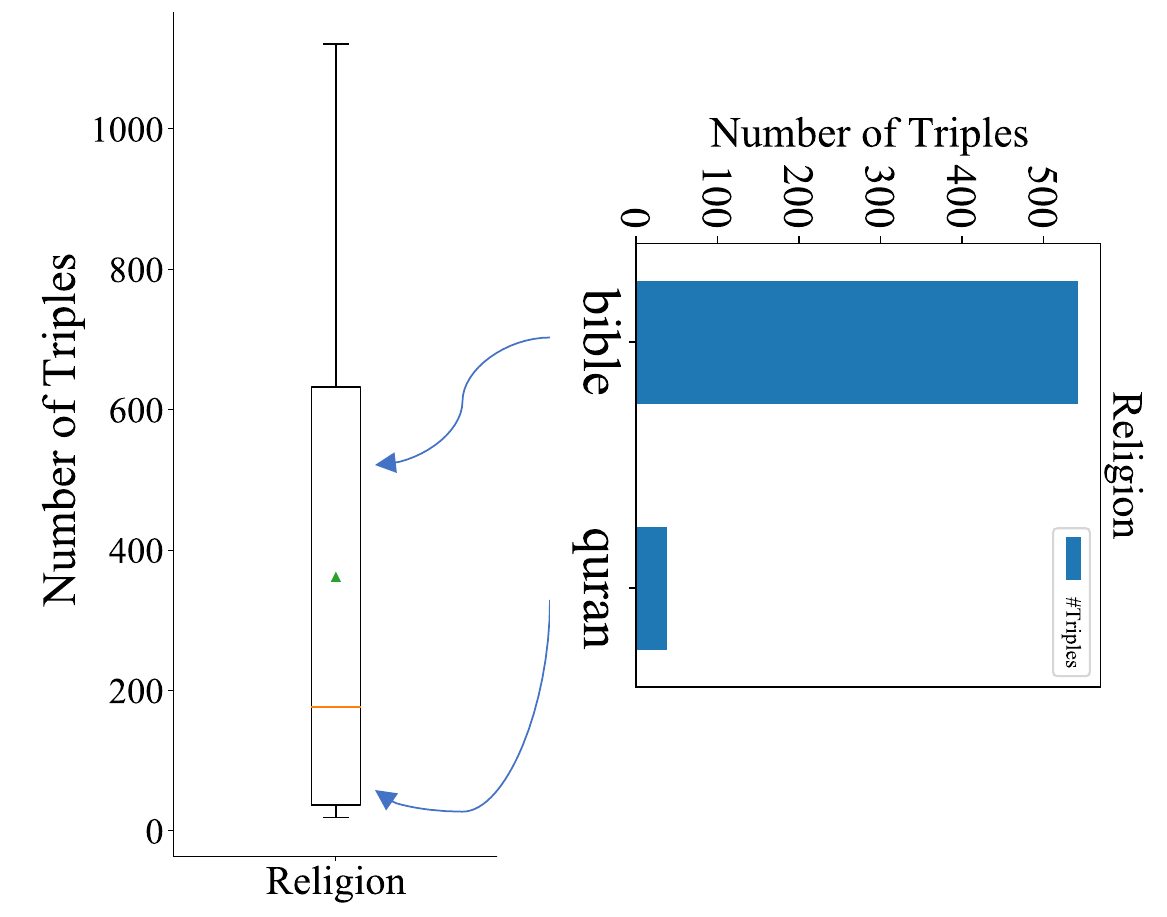}
\caption{ConceptNet Religion}
\end{subfigure}
\begin{subfigure}[b]{0.23\textwidth}
\includegraphics[width=\textwidth,trim=0cm 0.3cm 0cm 0cm,clip=true]{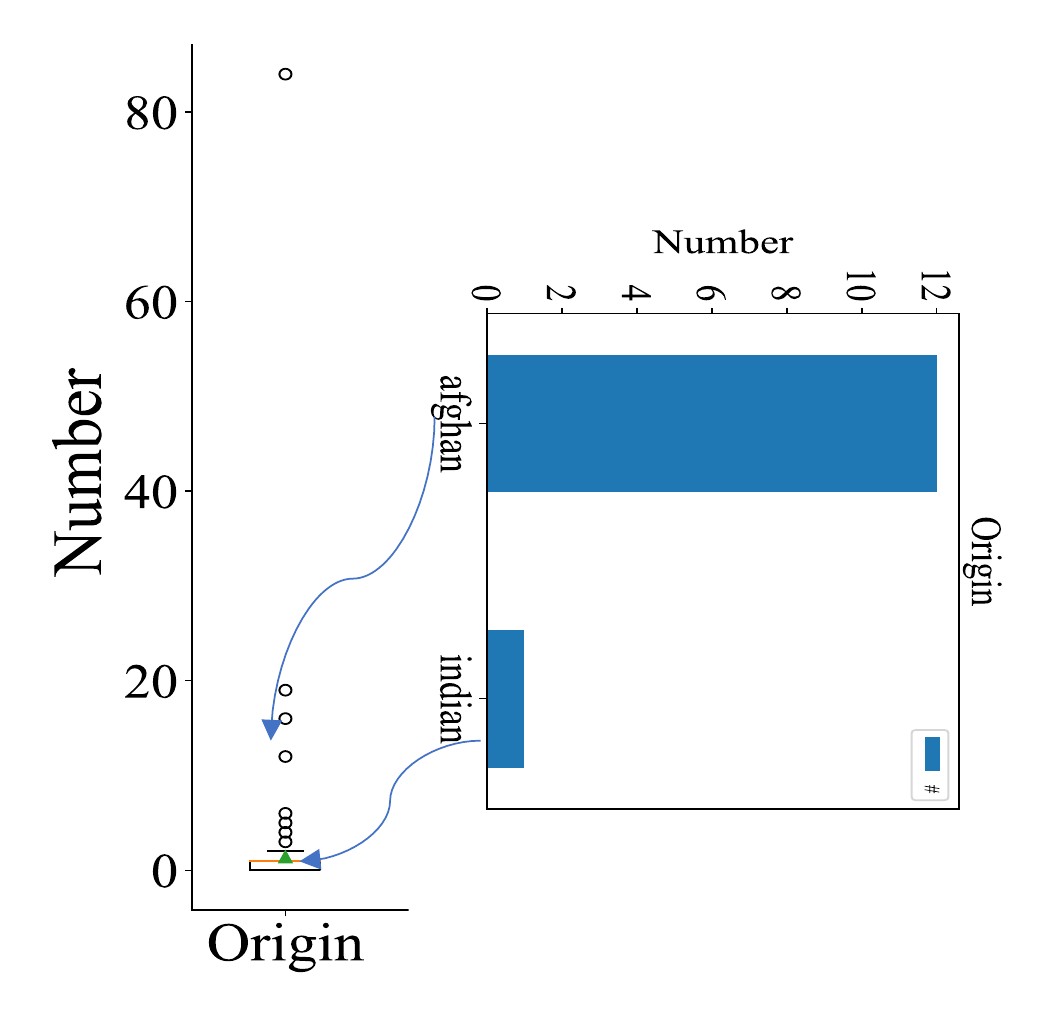}
\caption{GenericsKB Origin}
\end{subfigure}
\begin{subfigure}[b]{0.22\textwidth}
\includegraphics[width=\textwidth,trim=0cm 0.3cm 0cm 0cm,clip=true]{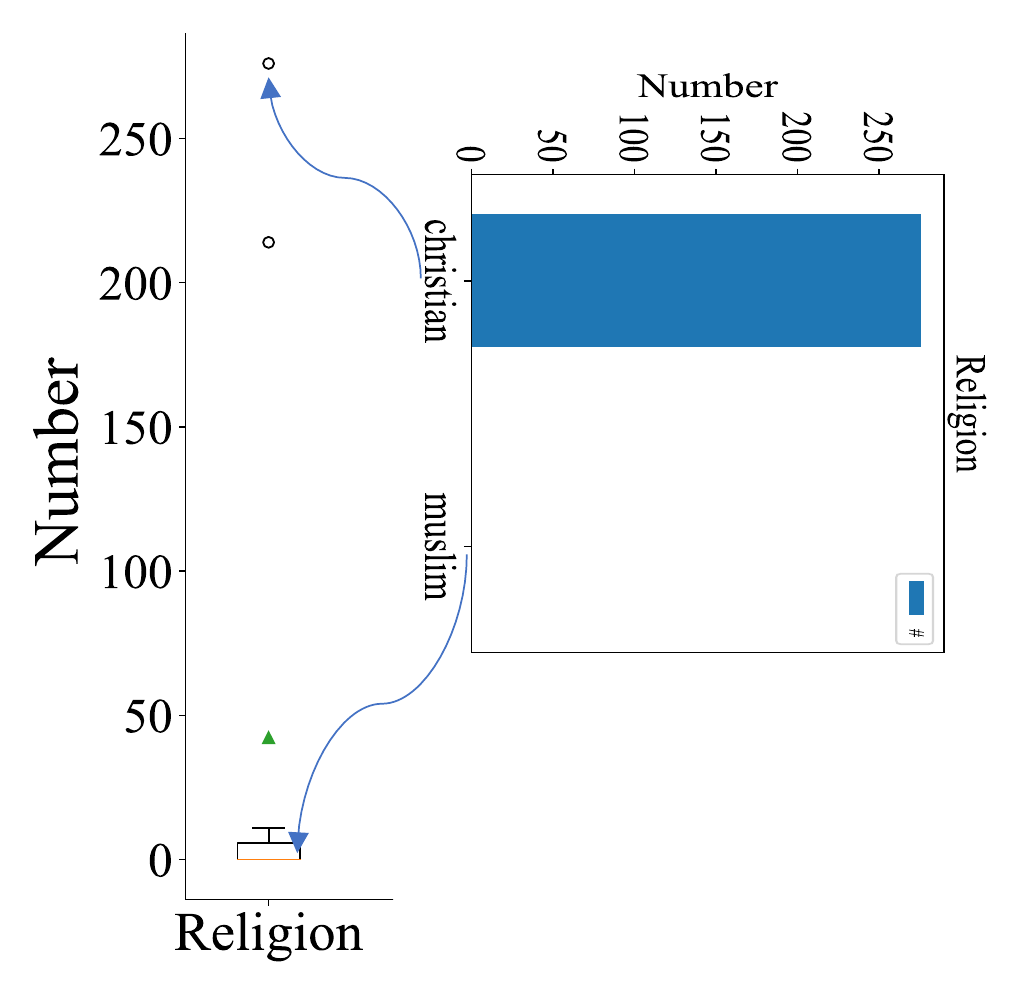}
\caption{GenericsKB Religion}
\end{subfigure}
\caption{Box plots demonstrating the \textbf{representation disparity} in terms of number of triples/sentences for \emph{Origin} and \emph{Religion} categories from ConceptNet and GenericsKB.}
\label{fig:remaining_freq}
\end{figure*}

\begin{figure*}[h]
\begin{subfigure}[b]{0.246\textwidth}
\includegraphics[width=\textwidth,trim=1.9cm 2.7cm 1.3cm 1.6cm,clip=true]{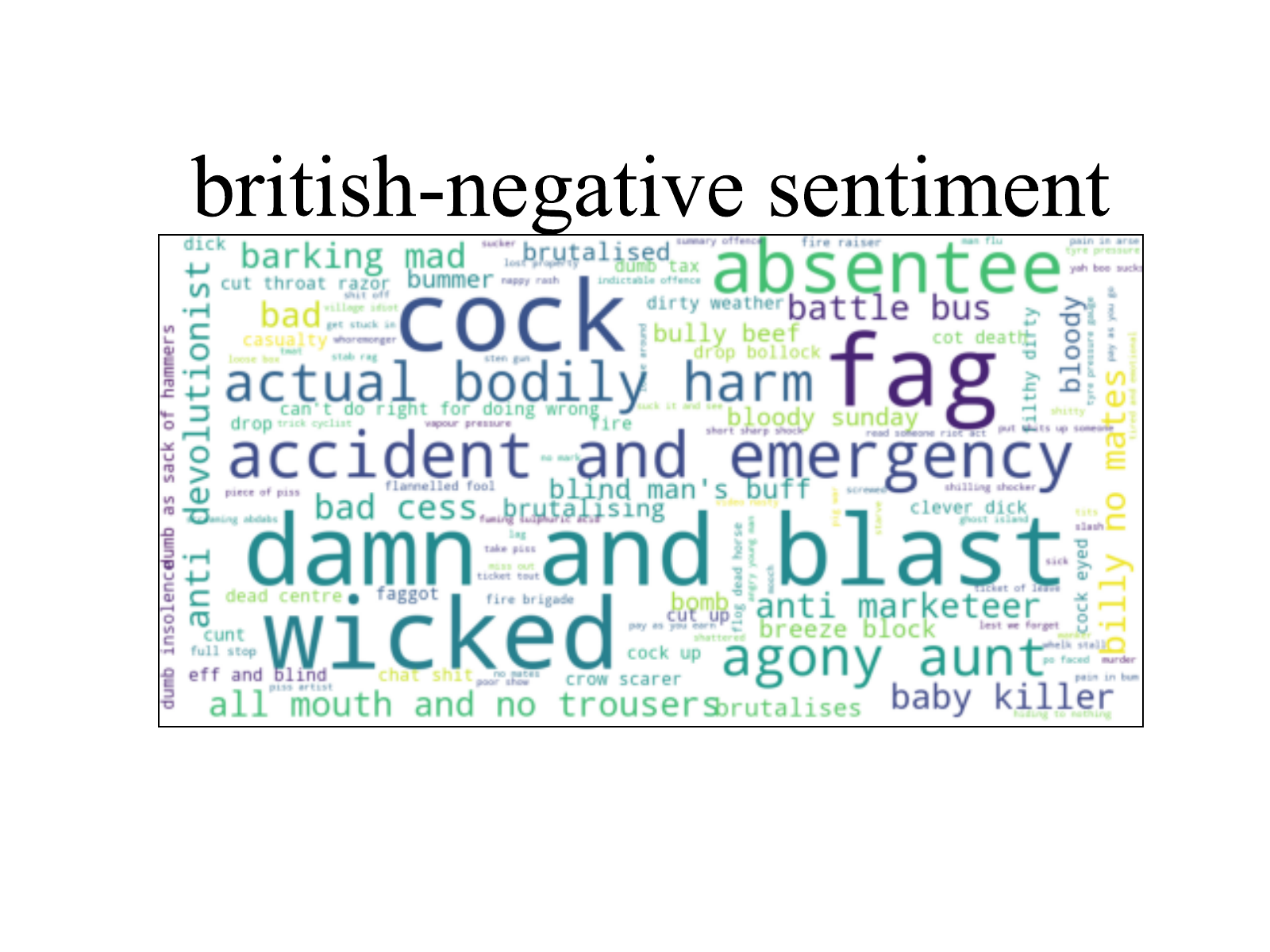}
\end{subfigure}
\begin{subfigure}[b]{0.246\textwidth}
\includegraphics[width=\textwidth,trim=1.9cm 2.7cm 1.3cm 1.6cm,clip=true]{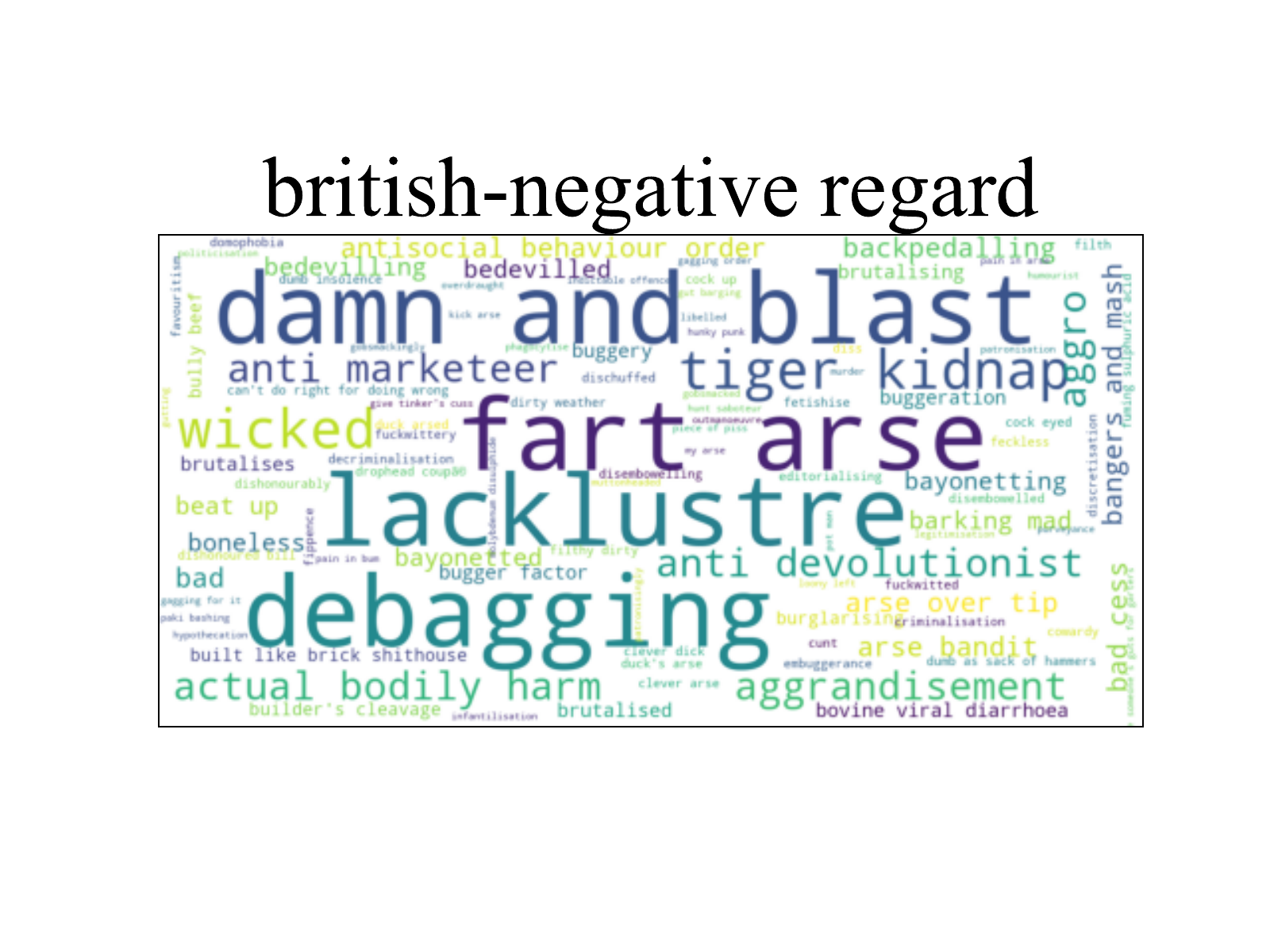}
\end{subfigure}
\begin{subfigure}[b]{0.246\textwidth}
\includegraphics[width=\textwidth,trim=1.7cm 2.7cm 1.3cm 1.6cm,clip=true]{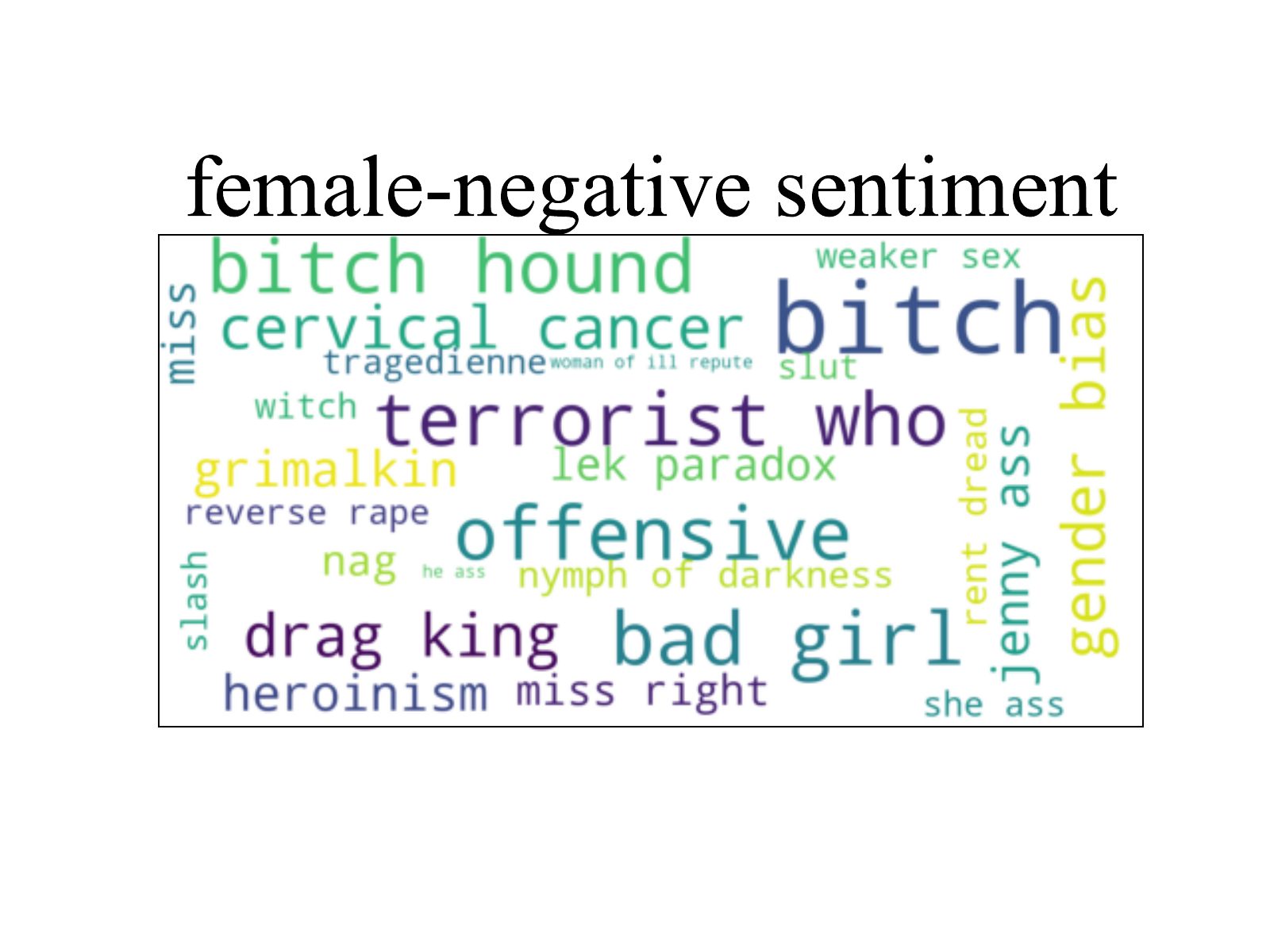}
\end{subfigure}
\begin{subfigure}[b]{0.246\textwidth}
\includegraphics[width=\textwidth,trim=1.9cm 2.7cm 1.3cm 1.6cm,clip=true]{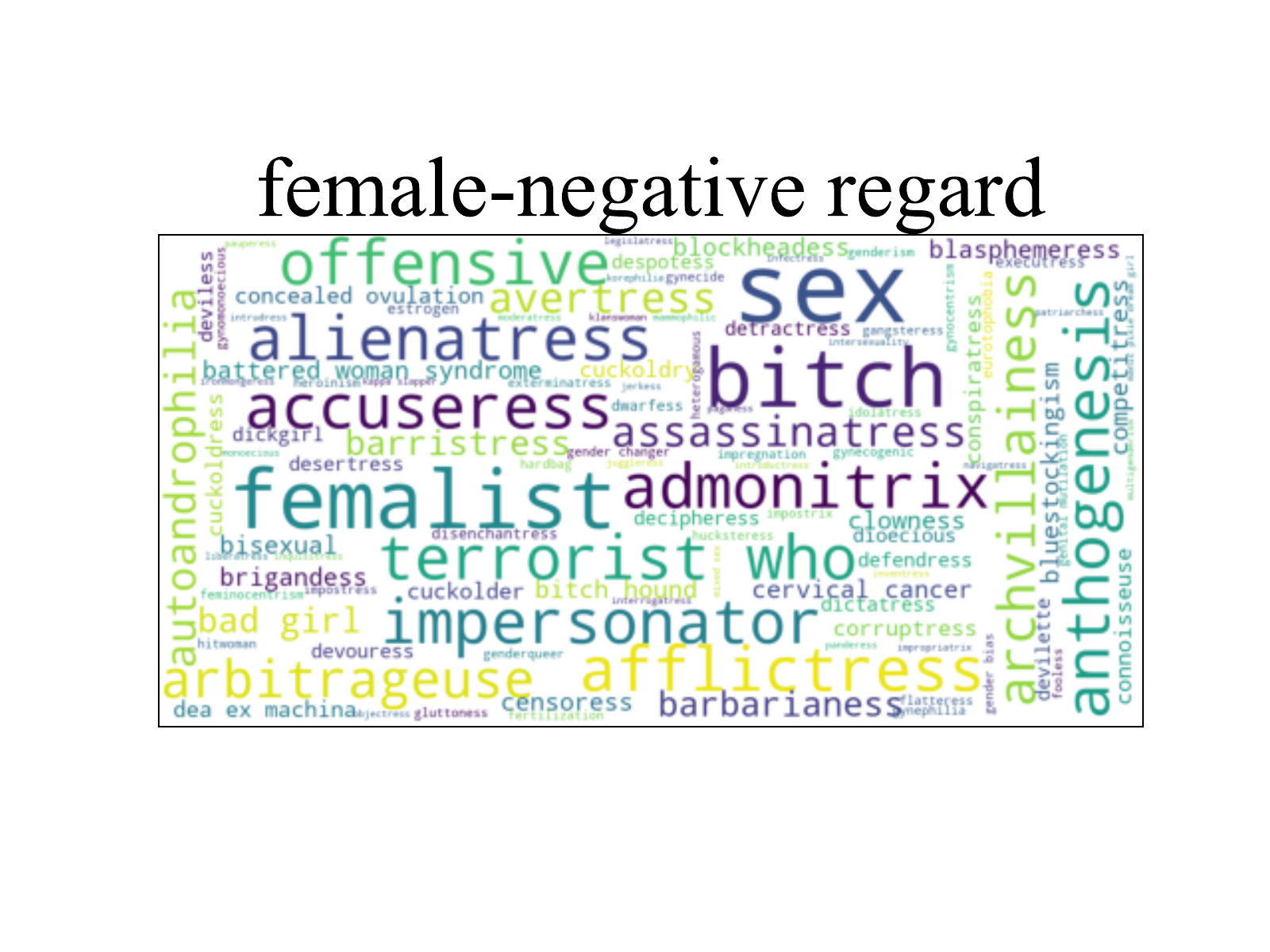}
\end{subfigure}
\caption{Wordcloud of phrases that appear in triples with negative regard and sentiment labels for ``\emph{british}'' and ``\emph{female}'' targets.}
\label{wordcloud_results_conceptnet}
\end{figure*}

\begin{table*}[h!]
\begin{tabular}{ p{2.3cm} p{3.3cm} p{2.3cm} p{2.8cm} p{3.0cm} }
 \toprule
 &&Profession&&\\
 \midrule
barber& coach& businessperson& football player& construction worker \\ manager&
CEO& accountant& commander& firefighter \\ mover& software developer&
guard& baker& doctor \\ athlete& artist& dancer&
mathematician& janitor \\ carpenter& mechanic& actor& handyman&
musician \\ detective& politician& entrepreneur& model& opera singer \\
chief& lawyer& farmer& writer& librarian \\ army&
real estate developer& broker& scientist& butcher \\ electrician& prosecutor&
banker& cook& hairdresser \\ prisoner& plumber& attorney&
boxer& chess player \\ priest& swimmer& tennis player& supervisor&
attendant \\ housekeeper& maid& producer& researcher& midwife \\
judge& umpire& bartender& economist& physicist \\ psychologist&
theologian& salesperson& physician& sheriff \\ cashier& assistant&
receptionist& editor& engineer \\ comedian& painter& civil servant&
diplomat& guitarist \\ linguist& poet& laborer& teacher&
delivery man \\ realtor& pilot& professor& chemist& historian \\
pensioner& performing artist& singer& secretary& auditor \\ counselor&
designer& soldier& journalist& dentist \\ analyst& nurse&
tailor& waiter& author \\ architect& academic& director&
illustrator& clerk \\ policeman& chef& photographer& drawer&
cleaner \\ pharmacist& pianist& composer& handball player& sociologist  \\[0.5pt]
  \bottomrule
\end{tabular}
\caption{Targets from the profession category in our dataset (most of them borrowed from Nadeem et al 2020). We considered triples that these words appeared in them.}
    \label{stereoset_prof_conceptnet}
\end{table*}

\begin{table*}[h!]
\begin{minipage}{.5\textwidth}
\begin{tabular}{ p{2.9cm} p{2.9cm} p{2.4cm} p{2.4cm} p{3cm}}
 \toprule
 &&Origin&\\
 \midrule
 African American&Armenian&Persian&American&Filipino \\English&Dutch&Israeli&Nigerian& Ethiopia \\ Europe& European& Russian& Ukraine&
Sudan \\ Afghanistan &Iraq& Yemen& Ukrainian& Russia \\
Italy& Somali& Iran& Afghan& Indian \\ Italian&
Australian& Spanish& Guatemalan& Hispanic \\ Venezuela& Sudanese&
Oman& Finnish& Swedish \\ Venezuelan& Puerto Rican& Ghanaian&
Moroccan& Somalia \\ Saudi Arabian& Syria& Chinese& Pakistani&
China \\ India& Irish& Britain& France& Greece \\
Scotland& Mexican& Paraguayan& Brazil& African \\ Eritrean&
Sierra Leonean& Africa& Jordan& Indonesia \\ Vietnam& Pakistan&
German& Romania &Brazilian \\ Ecuadorian& Mexico& Puerto Rico&
Kenyan& Liberian \\ Cameroonian& African Americans& Kenya& Liberia&
Sierra Leon \\ Qatari& Syrian& Arab& Saudi Arabia& Lebanon \\
Indonesian& French& Norwegian& South Africa& Jordanian \\ Korea
&Singapore& Romanian& Crimean& Native American \\ Germany& Ireland&
Ecuador& Morocco& Omani \\ Iranian& Iraqi& Qatar&
Turkey& Vietnamese \\ Nepali& Laos& Bangladesh& British&
Polish \\ Greek& Scottish& Bolivian& Guatemala& Ghana \\
Cameroon& Japanese& Taiwanese&Bengali& Nepal \\ Albanian&
Albania& Columbian& Peruvian& Argentian \\ Spain& Paraguay&
Ethiopian& Egyptian& Persian people \\ Sweden& Crimea& Portuguese&
Argentina& Chile \\ Cape Verdean& Turkish& Yemeni& Taiwan&
Austrian \\ White people& Finland& Australia& South African& Eriteria \\
Egypt& Korean& Dutch people& Peru& Poland \\ Chilean&
Columbia& Bolivia& Laotian& Lebanese \\ Japan& Norway&
Cape Verde& Portugal& Austria \\ Singaporean& Netherlands
 \\[0.5pt]
  \bottomrule
\end{tabular}
\\[0.5pt]
\end{minipage}%

\begin{minipage}{.45\textwidth}
\begin{tabular}{ p{2.9cm} p{2.9cm} p{2.4cm} p{2.4cm} p{3cm} }
 \toprule
 &&Gender&&\\
 \midrule
she&he&hers&him&her \\herself& himself& his&woman&man \\female&male&lady&gentleman&ladies \\gentlemen&girl&boy &sir& ma am \\mother&father&stepmother&stepfather&daughter \\ son&sister& brother&  grandmother&   grandfather  \\
  mommy& daddy&wife&husband&bride \\ groom&girlfriend& boyfriend&schoolgirl & schoolboy  \\[0.5pt]
  \bottomrule
\end{tabular}
\\[0.5pt]
\end{minipage}%

\begin{minipage}{.45\textwidth}
\begin{tabular}{ p{2.9cm} p{2.9cm} p{2.4cm} p{2.4cm} p{3cm} }
 \toprule
 &&Religion&&\\
 \midrule
Sharia& Jihad& Christian& Muslim& Islam \\ Hindu&
Mohammed& church&  Quran&Bible \\Brahmin&Holy Trinity  \\[0.5pt]
  \bottomrule
\end{tabular}
\\[0.5pt]
\end{minipage}%

\caption{Targets from origin, gender, and religion categories in our dataset (most of them borrowed from Nadeem et al 2020). We considered triples that these words appeared in them.}
    \label{stereoset_groups_conceptnet}
\end{table*}

\end{document}